\documentclass{article} 
\usepackage{iclr2026_conference,times}

\usepackage{amsmath,amsfonts,bm}

\def\eqref#1{equation~\ref{#1}}

\def\1{\bm{1}}

\DeclareMathAlphabet{\mathsfit}{\encodingdefault}{\sfdefault}{m}{sl}
\SetMathAlphabet{\mathsfit}{bold}{\encodingdefault}{\sfdefault}{bx}{n}

\iclrfinalcopy

\usepackage{hyperref}
\usepackage{url}
\usepackage{amssymb}
\usepackage{pifont}
\usepackage{graphicx}
\usepackage{booktabs}
\usepackage{xcolor}
\usepackage{multirow}
\usepackage{makecell}
\usepackage{subcaption}
\usepackage{float}
\usepackage{wrapfig}
\usepackage{listings}
\usepackage{enumitem}

\title{TRAJECT-Bench:A Trajectory-Aware Benchmark for Evaluating Agentic Tool Use}

\author{Pengfei He$^1$, Zhenwei Dai$^2$, Bing He$^2$, Hui Liu$^2$, Xianfeng Tang$^2$, Hanqing Lu$^2$, Juanhui Li$^2$,\\
\textbf{Jiayuan Ding$^3$, Subhabrata Mukherjee$^3$, Suhang Wang$^4$, Yue Xing$^1$, Jiliang Tang$^1$, Benoit Dumoulin$^2$}\\
$^1$Michigan State University~~~$^2$Amazon Inc.~~~$^3$Hippocratic AI~~~$^4$Penn State University\\
\\
\raisebox{-0.35\height}{\includegraphics[scale=0.045]{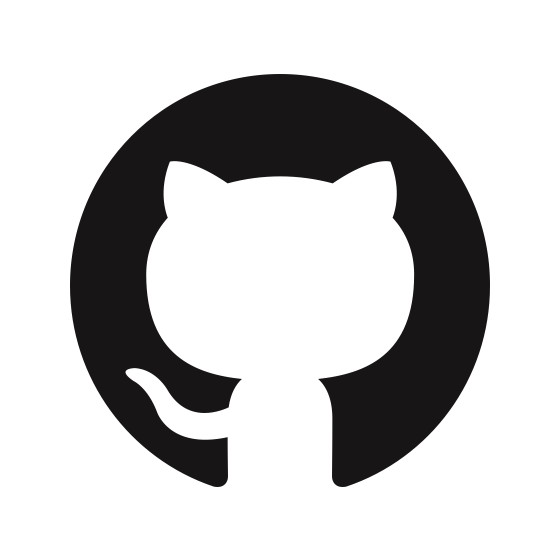}} \textbf{Github: \href{https://github.com/PengfeiHePower/TRAJECT-Bench}{https://github.com/PengfeiHePower/TRAJECT-Bench}} \\
\raisebox{-0.35\height}{\includegraphics[scale=0.1]{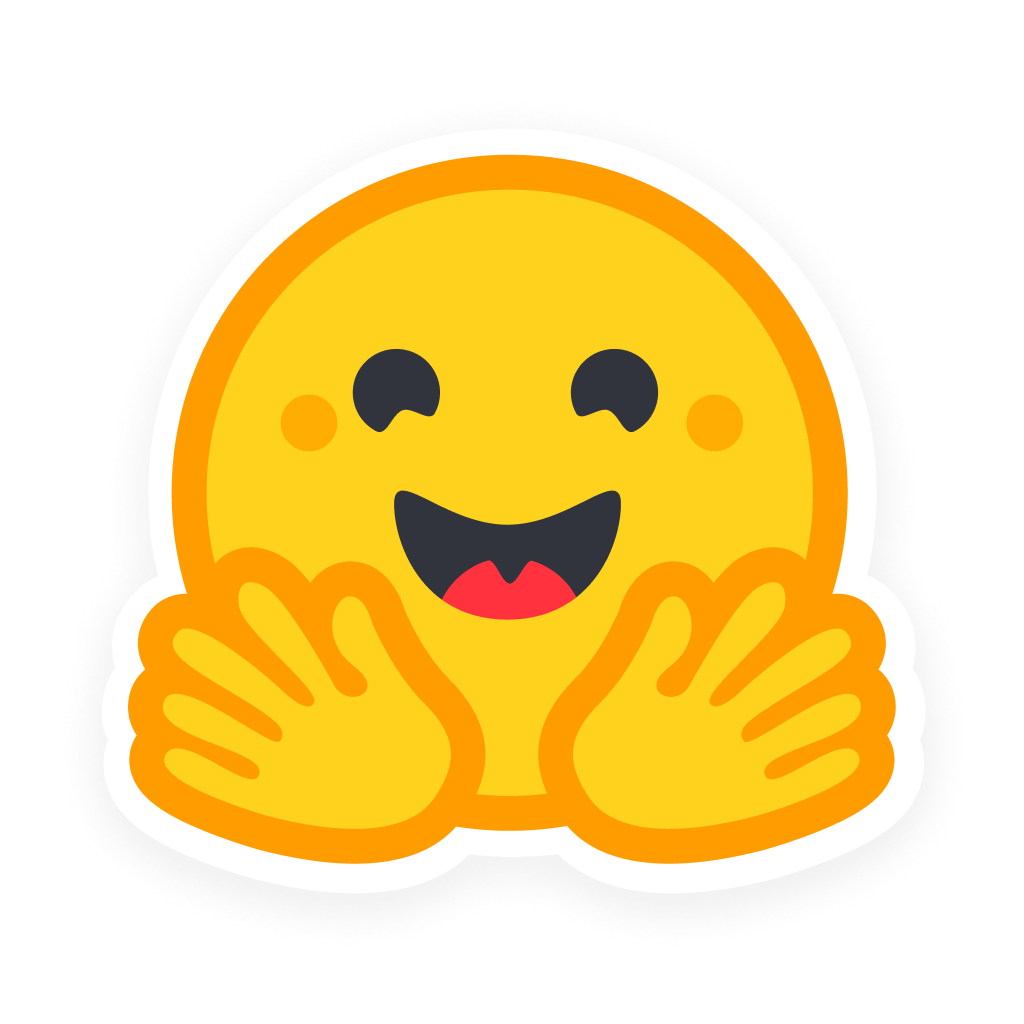}} \textbf{Higgingface: \href{https://huggingface.co/datasets/bigboss24/TRAJECT-Bench}{https://huggingface.co/datasets/bigboss24/TRAJECT-Bench}}
}

\newcommand{\greentick}{\textcolor{green}{\ding{51}}}
\newcommand{\redcross}{\textcolor{red}{\ding{55}}}
\definecolor{MySteelBlue}{HTML}{4682B4}
\newcommand{\bench}{\textcolor{MySteelBlue}{{\texttt{TRAJECT-Bench}}}}

\begin{document}

\maketitle

\begin{abstract}

Large language model (LLM)-based agents increasingly rely on tool use to complete real-world tasks. While existing works evaluate the LLMs' tool use capability, they largely focus on the final answers yet overlook the detailed tool usage trajectory, i.e., whether tools are selected, parameterized, and ordered correctly. We introduce TRAJECT-Bench, a trajectory-aware benchmark to comprehensively evaluate LLMs' tool use capability through diverse tasks with fine-grained evaluation metrics. TRAJECT-Bench pairs high-fidelity, executable tools across practical domains with tasks grounded in production-style APIs, and synthesizes trajectories that vary in breadth (parallel calls) and depth (interdependent chains). Besides final accuracy, TRAJECT-Bench also reports trajectory-level diagnostics, including tool selection and argument correctness, and dependency/order satisfaction. Analyses reveal failure modes such as similar tool confusion and parameter-blind selection, and scaling behavior with tool diversity and trajectory length where the bottleneck of transiting from short to mid-length trajectories is revealed, offering actionable guidance for LLMs' tool use.
\end{abstract}

\section{Introduction}
Large language models (LLMs) increasingly exhibit strong planning and reasoning abilities \citep{wei2022chain, yao2023tree}: they break goals into subproblems, track intermediate states, and arrange multi‑step procedures \citep{huang2024understanding, song2023llm}. Together, these abilities position LLMs as the ``brain" of agentic systems, guiding the systems to perform complex, real‑world tasks \citep{wang2024survey, he2024make, li2024survey}. Meanwhile, external tools, such as search engines \citep{jin2025search}, production APIs \citep{li2023api}, and file/OS operations \citep{packer2023memgpt}, function as the agent’s ``hands", extending LLM's abilities with precise computations, up‑to‑date information, and concrete actions. Tool-use has already powered agents across domains: travel agents \citep{chen2024travelagent, singh2024personal} fuse flight and hotel APIs with visa and weather services to assemble feasible itineraries and resolve constraints; and education agents \citep{chu2025llm, zhang2024simulating} retrieve curriculum materials, generate adaptive exercises, grade against rubrics, and track progress across learning platforms. 

Despite these developments, proper evaluations of LLMs’ tool-use are essential: they can provide a complete picture of tool‑use competence, expose failure modes, and point out concrete directions for improvement. Specifically, the capability to select the right tool(s), determine the correct formats and values of input parameters, conduct multi-step tool utilization, and adapt to unseen tools, makes the core part of evaluating tool-use capability. Benchmarks have been created to test these abilities. For example, \citet{huang2023metatool} focus on evaluating whether LLMs can determine when to call tools, and \citet{qin2023toolllm, patil2024gorilla, patilberkeley, zhuang2023toolqa} evaluate LLMs' capability in calling proper tools for solving complex queries.

However, significant gaps in tool-use evaluation still persist. First, tool-use trajectory complexity is comparatively underexplored: some existing suites rely on small or simulated tools \citep{zhuang2023toolqa} and many only test short, low‑depth tool trajectories~\citep{qin2023toolllm}. However, real agents can possess large tool sets and face complex user queries, so evaluation on a larger executable tool set and trajectories with more tools involved is still needed. Second, the complexity of real user queries is also underrepresented: existing benchmarks often consider straightforward prompts by including the API name directly in the prompt. In contrast, real-world agents may face user queries that consist of indirect language and implicit cues, and the agents need to infer both the choice of the tool and how to set the inputs. Third, most benchmarks still privilege final‑answer metrics. For example, \citet{qin2023toolllm} only provide pass rate and win rate of the final answer, and \citet{patilberkeley} heavily rely on the overall accuracy. In this case, it is hard to track the root cause of the incorrect final answer, which can be caused by various issues such as incorrect tool selection, disorders of tool-use, or incorrect parameterization. This oversight can obscure the evaluation of LLMs' tool-use capability and cannot differentiate it from other general reasoning capabilities, especially given the observation in \citep{roberts2020much, qian2025smart} that LLMs can solve the problem using internal knowledge even when the wrong tools are called.

\begin{table}[t]
\centering
\caption{Comparison of previous work and \bench.}
\vspace{-10pt}
\label{tab:compare}
\resizebox{\columnwidth}{!}{%
\begin{tabular}{lcccccccc}
\toprule
                       & \makecell{\textbf{Practical}\\ \textbf{tools}} & \makecell{\textbf{Large\&diverse}\\ \textbf{tool}} & \makecell{\textbf{Trajectory}\\ \textbf{structure}} & \makecell{\textbf{Trajectory}\\ \textbf{scaling}} & \makecell{\textbf{Trajectory-aware}\\ \textbf{metrics}} & \makecell{\textbf{Query}\\ \textbf{difficulty}} & \makecell{\textbf{Tool}\\ \textbf{selection}} & \makecell{\textbf{Agentic}\\ \textbf{evaluation}} \\
                       \midrule
\textbf{MetaTool \citep{huang2023metatool}}                         & \greentick               & \redcross                   & \redcross                     & \redcross                   & \redcross                         & \redcross                        & \redcross                  & \redcross               \\
\textbf{API-Bank \citep{li2023api}}                         & \greentick               & \redcross                   & \redcross                     & \redcross                   & \redcross                         & \redcross                        & \redcross                  & \redcross               \\
\textbf{ToolBench \citep{qin2023toolllm}}                        & \greentick               & \greentick                   & \redcross                     & \redcross                   & \redcross                         & \redcross                        & \redcross                  & \greentick               \\
\textbf{Gorilla \citep{patil2024gorilla}}                          & \greentick               & \greentick                   & \redcross                     & \redcross                   & \redcross                         & \redcross                        & \greentick                  & \redcross               \\
\textbf{BFCL \citep{patilberkeley}} & \greentick               & \greentick                   & \redcross                     & \redcross                   & \redcross                         & \redcross                        & \redcross                  & \redcross               \\
\textbf{ToolQA \citep{zhuang2023toolqa}}                           & \redcross               & \redcross                   & \redcross                     & \redcross                   & \redcross                         & \greentick                        & \redcross                  & \greentick               \\
\textbf{TRAJECT-Bench} (ours)         & \greentick           & \greentick               & \greentick                 & \greentick               & \greentick                     & \greentick                    & \greentick              & \greentick \\
\bottomrule
\end{tabular}
}
\end{table}
 
To bridge these gaps, we present \bench--a benchmark to comprehensively evaluate the LLMs tool-use capability via providing \textbf{(1) tool-use trajectories} of different complexities and \textbf{(2) user queries of different levels of difficulties} given the same tool-use trajectory. Meanwhile, we also include \textbf{(3) evaluation metrics} to evaluate the LLMs' tool-use capability from diverse perspectives. As summarized in Table~\ref{tab:compare}, we are the first to provide all the comprehensive analysis perspectives for tool-use evaluation compared to existing benchmarks.

During the data construction, to ensure the quality of the data and better align it with real scenarios, on the tool side, we select a diverse suite of over 1,000 high-fidelity tools drawn from various real-world domains (e.g., finance, travel, music, etc.), exposed via production-style APIs. Given these tools, to construct \textbf{(1) tool-use trajectories}, we further synthesize task-driven tool-use trajectories that encode different trajectory structures and scales. We consider both parallel and sequential tool calling trajectory structures and a trajectory scale of 3 to 10+ tool counts. Moreover, for each trajectory, we provide \textbf{(2) two semantically aligned queries of different query difficulties}: a direct and explicit ``simple'' version and a naturalistic and indirect ``hard'' version. Such a data construction procedure allows us to decompose the queries' difficulty into different perspectives and understand the corresponding specific weaknesses of LLMs. 

Based on this data, we evaluate the state-of-the-art models, compare representative tool-selection strategies, and assess agentic tool-use settings beyond individual LLMs. For \textbf{(3) the evaluation metrics}, we report trajectory-aware metrics alongside final-answer accuracy: Trajectory Exact-Match and Trajectory Inclusion (whether the required tools are invoked and in the correct order), Tool-Usage (schema constraints, formats, and value checks of the tool's inputs), and an LLM-judge Trajectory-Satisfy score when gold traces are unavailable. Together, these evaluations treat tool-use as a primary skill, with trajectory-aware metrics that explicitly measure trajectory and query complexity. Our key contributions are summarized as follows:
\vspace{-5pt}
\begin{itemize}[leftmargin=1.5em] 
\item \textbf{Dataset and tasks.} We introduce high-quality, executable tool suites across various domains and tasks. We model trajectory complexities via different structures (\emph{parallel} and \emph{sequential}) and various tool counts. Queries are aligned with trajectories with different difficulty levels.
\vspace{-3pt}
\item \textbf{Evaluation suite.} We evaluate state-of-the-art LLMs and representative tool selection strategies. We also include agentic tool-use in the evaluation suite. Beyond final-task accuracy, we introduce trajectory-aware metrics to better capture tool-use capability.
\vspace{-3pt}
\item \textbf{Insights and guidance.} We provide analyses based on our evaluation, and reveal important insights, such as models' struggling to infer correct tools from indirect queries, the bottleneck of the transition from short to mid-length trajectories, limitations of retrieval-based selection. These insights can inspire future development of a more precise and reliable tool-use mechanism.
\end{itemize}

\section{Related Work}
\textbf{LLM Agents}.
LLM-based agents treat large language models as the “brain” that plans, acts, and reflects while interfacing with external tools and environments \citep{wang2024survey, guo2024large}. Early formulations follow a think–act loop: ReAct interleaves chain-of-thought with grounded actions \citep{yao2023react, wei2022chain}, and Reflexion adds self-critique and memory to correct future behavior \citep{shinn2023reflexion}. Building on this, tool-augmented frameworks formalize function calling, tool selection, and state management to operate retrieval, code execution, and other systems \citep{schick2023toolformer, lewis2020retrieval}. These designs have made LLM agents effective and flexible across domains \citep{he2024make, trivedi2022interleaving, li2023chain}. Web agents perceive pages and perform UI-level actions for tasks like shopping and booking \citep{yao2022webshop, zhou2023webarena, he2024webvoyager, zheng2024gpt}; code agents iteratively edit, test, and debug multi-file projects \citep{yang2024swe, hong2024metagpt}; embodied agents couple language with perception and control \citep{kim2024understanding, mao2023language, zhang2024towards}; and scientific agents use tools and simulation for literature review, hypothesis generation, and experiment planning \citep{10.1145/3586183.3606763, ren2025towards}.

\textbf{Tool-use in LLM agents}. Tool use is a core capability of LLM agents and has drawn substantial interest \citep{qu2025tool, qin2024tool}. Methods advance when and how models invoke functions: RL approaches (ReTool, ToolRL, OTC) optimize selection and calling \citep{feng2025retool, qian2025toolrl, wang2025otc}; supervised/instruction tuning (e.g., Toolformer; small-model instruction tuning) teaches adherence to schemas \citep{schick2023toolformer, shen2024small}; and feedback frameworks (TRICE) add self-correction \citep{qiao2023making}. Complementary evaluations measure these abilities: Gorilla grounds calls in public APIs \citep{patil2024gorilla}; the Berkeley Function-Calling Leaderboard scores cross-domain execution \citep{patilberkeley}; ToolBench scales to multi-step RapidAPI tools \citep{qin2023toolllm}; ToolQA targets tool-augmented reasoning \citep{zhuang2023toolqa}; and MetaTool probes tool awareness and selection \citep{huang2023metatool}. Together, this work advances both capability and measurement for tool-using agents. However, existing benchmarks seldom treat tool use as the primary objective. Most prioritize end‑task scores and overlook trajectory‑level signals, limiting the insight into agent behavior and opportunities for improvement. Therefore, a benchmark with a high‑fidelity, diverse tool set and realistic, task‑driven queries spanning difficulty levels, enabling trajectory‑aware evaluation and diagnosis, is still urgently needed.
\section{Data Construction}
To construct the data, we first select the candidate tools to be included in the benchmark (Section \ref{sec:data:tool}), and then curate ground truth tool-using trajectories and corresponding queries (Section \ref{sec:data:query}).
\vspace{-0.05in}
\subsection{tool set Curation}\label{sec:data:tool}
\vspace{-0.05in}
Following \citep{qin2023toolllm}, we source practical tools from RapidAPI, which contains many APIs for real-world tasks. To keep the effort tractable, we curate data for ten representative domains: travel, mapping, finance, weather, e-commerce, news/media, gaming, email, education, and music, where LLM agents are especially popular and useful\footnote{More domains can be added following our data-generation pipeline}. 

Although thousands of APIs are available from RapidAPI, many of them contain unclear descriptions or parameters, and some are even non-executable. Therefore, we carefully select and revise the APIs based on the following four requirements.

(1) \textbf{Executable tools with meaningful outputs}. We validate each tool by executing it across parameter combinations and discard those that error out (details in the Appendix \ref{app:tool details}). For the remainder, we use an LLM to summarize outputs and formats, removing tools whose outputs are semantically trivial, that is, do not contribute useful information. 

(2) \textbf{Clear, action-oriented tool descriptions}. Because many APIs ship with sparse or vague documentation, we refine API descriptions by combining the original description with empirical I/O observed during validation. For instance, the original description is just ‘Get price (symbol),’ yet empirical calls reveal it enforces a 50‑item page\_size cap and returns \{price, currency, timestamp\}; we merge these behaviors into the clarified description. 

(3) \textbf{Minimal functional overlap}. We deduplicate identical tools/APIs (e.g., multiple flight-search endpoints) to avoid ambiguity and ensure deterministic trajectory evaluation, keeping a single representative per function. Closely related tools are kept when they introduce distinct parameterizations to increase task complexity. This is done with the help of LLMs, followed by manual verification.

(4) \textbf{Controlled tool complexity}. We manually keep tools with strong parameter complexity (number of fields, types, and constraints) and remove some with simpler parameters (e.g., no inputs needed), ensuring the set contains rich tools to stress-test tool-use competence. 
Finally, we obtain a high-fidelity tool set $\mathcal{T}$. Details and examples of the curated tool set can be found in the Appendix \ref{app:tool details}.

\begin{figure}[t]
    \centering
    \includegraphics[width=\textwidth]{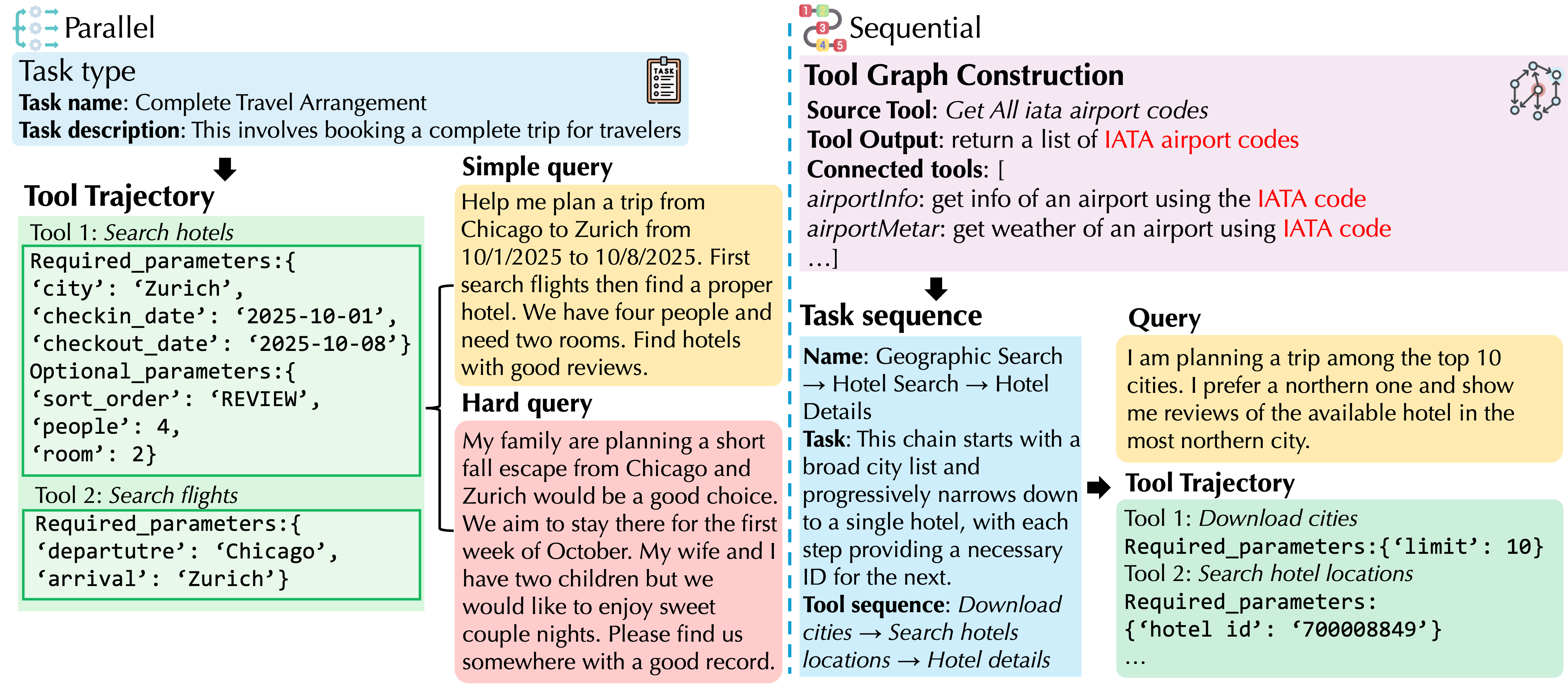}
    \vspace{-18pt}
    \caption{\small{An illustration of data in \bench. The left side demonstrates the parallel query where tool trajectories are created based on real task types and then queries with two difficulty level are generated. The right side shows the generation process of sequential queries, where a tool graph is first built, then task sequences are manually designed and finally detailed queries and trajectories are created.}}
    \label{fig:illustrate}
\end{figure}
\vspace{-0.05in}
\subsection{Task-driven Query Generation}\label{sec:data:query}
\vspace{-0.05in}
To evaluate LLMs' tool use in realistic settings while keeping the evaluation controllable, we synthesize queries from real-world \textit{task types}, which denote categories/families of tasks that users can adopt the agent for, such as booking flights and arranging stays in the destination (without details of dates or locations). 
Specifically, as shown in the left side of Figure \ref{fig:illustrate}, for each domain we manually collect representative task types (e.g., "real-time trip monitoring \& assistance" for travel; "creating math \& science study materials" for education), and provide a precise description for each task type.
For every task type, we create tool-use trajectories with details and pair them with aligned user queries. By templating from task types, we minimize human efforts in collecting and annotating data, yet keep queries faithful to real use cases. This also enables diverse, controllable, and scalable query–trajectory data across domains.

We consider two basic trajectory structures: (i) \textit{parallel}, where tools operate independently (the choice/execution of one tool does not depend on others); and (ii) \textit{sequential}, where tools form a chain and later steps depend on earlier outputs\footnote{We focus on chain structures to balance data cost and evaluation reliability; richer graph topologies are left for future work.}. Details of trajectory and query construction are provided as follows.
\vspace{-0.05in}
\subsubsection{Parallel query}\vspace{-0.05in}
\textbf{Trajectory generation}. 
For each domain, we prompt an LLM to synthesize valid, logical tool-using trajectories from a task-type description and the domain’s available tools. To keep evaluation scalable and comparable, we enforce two rules: (i) the plan must use a specified number of tools (typically 3–10+); and (ii) tools are presented as a parallel-ready set—each call is self-contained (inputs fixed up front) and independent of the others. Parallel trajectories are encoded as unordered sets of tool calls with fully specified inputs. For example, Figure~\ref{fig:illustrate} shows that combining hotel and flight search APIs enables end-to-end travel planning.

\textbf{Query generation}. After trajectories are finalized, we pair each trajectory with user queries designed at two difficulty levels: 
\vspace{-8pt}
\begin{itemize} 
\item \textbf{Simple version}. Provide a straightforward and precise instruction that explicitly specifies the need for tools and their key parameters, such as the example in Figure \ref{fig:illustrate} where the simple query gives direct and detailed requests for a travel plan.
\item \textbf{Hard version}. Present a more challenging, indirect request that conveys the same constraints via natural cues and implications rather than explicit instructions. This mirrors real interactions where users state goals colloquially (e.g., \emph{"hotels with good record"} rather than \emph{"sort hotels by review score"} as shown in Figure \ref{fig:illustrate}). 
\end{itemize}
\vspace{-8pt}
After the generation, both trajectories and queries go over LLM-based automatic validation and human inspection to ensure the quality and reduce ambiguity. By jointly varying (i) the number of tools per trajectory and (ii) the difficulty level of paired queries, we obtain a comprehensive yet well-controlled evaluation of LLMs’ tool-use capabilities.
\vspace{-0.05in}
\subsubsection{Sequential query}
\vspace{-0.05in}
Sequential queries require strong dependencies between tools within the trajectory, which makes it difficult for an LLM to directly propose a correct chain end-to-end. To address this, we first build a \textbf{directed tool graph} $G_T=(V,E)$ that captures how tools can feed one another: each tool $t$ in $V$ is a node, and we add a directed edge $t_1 \rightarrow t_2$ to the edge set E, when the information from the output of $t_1$ can be used as input parameters of $t_2$. For instance, as shown in the right side of Figure \ref{fig:illustrate}, the IATA codes returned from the GetAllIATA API can be used as the input for airportInfo API to retrieve detailed information of a specific airport, so we connect them.

Given $G_T$, we generate representative tasks and corresponding tool sequences with a controlled number of tools (e.g., 5 sequences for each number of tools). Unlike the parallel setting, these sequences are manually created as trajectory templates to ensure logical coherence, e.g., $t_1 \rightarrow t_2 \rightarrow \cdots \rightarrow t_{n_traj}$, where $n_traj$ denotes the number of tools involved. Each template explicitly specifies the parameter bindings between adjacent tools, i.e., which output results from $t_i$ populate which input fields of $t_{i+1}$. For instance, the task sequence in Figure \ref{fig:illustrate} describes a general 'city->hotel location->hotel detail' chain without details such as city name or hotel id. 

Similar to the generation of parallel trajectories, we then prompt an LLM to create trajectories with details and corresponding user queries from each template (5 per template), yielding $s^{seq}_{traj} = \{t_1(params) \rightarrow \cdots \rightarrow t_{n_{traj}}(params)\}$ together with $q^{seq}_{traj}$, as shown in Figure \ref{fig:illustrate}. All trajectories and their paired queries undergo automated validation and human review to ensure validity, task alignment, and overall quality. More details of the generation process and datasets are shown in the Appendix \ref{app:traj details}. This pipeline produces coherent tool chains and tightly aligned queries at varied depths (number of tools involved), enabling scalable and transparent evaluation.

Finally, the whole dataset contains 1228 tools and 5670 queries (details in Appendix \ref{app:traj details}).


\section{Evaluations on \bench}
We comprehensively evaluate LLMs’ agentic tool-use on \bench, focusing on key research questions that highlight its benefits and potential:

RQ1. When and why do LLMs succeed or fail at tool-use?

\vspace{-4pt}
RQ2. Can retrieval-augmented selection improve tool selection and parametrization?

\vspace{-4pt}
RQ3. Do agentic methods, including training and inference, improve tool-use capability?
\vspace{-10pt}
\subsection{Evaluation settings}\label{sec:eval setting}
\vspace{-5pt}
\textbf{Test (query) methods.} When evaluating LLM's own capability in calling proper tools to solve queries, we mainly adopt the direct query where user queries and tools are provided to the model. We provide additional results using Chain-of Thought in Appendix \ref{app:exp}.

\textbf{Tool selection strategy.} Unlike existing benchmarks \citep{qin2023toolllm, huang2023metatool} ship only retrieved tools (not the full tool set) during evaluation, we treat the tool selection strategy as a crucial part of tool-using, especially when the tool set is large and exceeds the model’s context window. We evaluate three common tool selection strategies (labels shown in Table \ref{tab:retrieval,parallel}): all—provide the full tool set in context; domain—provide only tools from the query’s domain; and retrieval—retrieve a subset based on the query and tool descriptions. For retrieval, we test two widely used embedding models, all-MiniLM-L6-v2 \citep{solatorio2024gistembed} and bge-large-en-v1.5 \citep{bge_embedding}, plus a tool-specific retriever, ToolBench-IR \citep{qin2023toolllm}. Unless noted, we retrieve 20 tools by default.

\textbf{Test models.} We test on a wide range of state-of-the-art LLMs: Claude family, including Claude-3.7 \citep{anthropic2024_claude3} and Claude-4 \citep{anthropic2025_claude4_systemcard}; Gemini family, including Gemini-2.5-pro and Gemini-2.5-flash; GPT family, including o4-mini and gpt-oss-120B; Qwen3-235b-A22B, the latest generation of LLMs in the Qwen series; DeepSeek-V3.1, the latest DeepSeek model; and Kimi-k2, the latest MoE model from MoonShot AI. 

\textbf{Agentic evaluation.} Despite the tool-using capability of models themselves, one highlight of our benchmark is the evaluation of models' agentic tool-using capability. We consider two agentic capabilities. The first one is the capability specifically trained in the models. The SOTA LLMs, such as Claude, Gemini, Deepseek, and Kimi-k2, all include tool-using in the training and directly support tool-calling, e.g., the ``tools" input in \texttt{client.messages.create} of Claude APIs, and do not need to additionally list them in the context of the prompt. Therefore, we provide evaluations on these models' internal tool-using capability and compare with the results of providing tools as context. More details can be found in Appendix \ref{app: exp detail}.

We also evaluate LLM agents for tool-use. Given diverse designs, we focus on ReAct \citep{yao2023react}—a foundational blueprint for many real agents (e.g., MetaGPT \citep{hong2024metagpt}). We pair ReAct with domain-specific and retrieval-based tool selection. Because ReAct reasons and executes over multiple turns, we test two retrieval modes: \emph{static} (retrieve once from the query) and \emph{dynamic} (retrieve at each reasoning/execution turn).

\textbf{Metrics.} Unlike existing benchmarks, which mainly focus on the final performance, we consider two categories of metrics: trajectory-aware metrics and final performance metrics. For the trajectory-aware metrics, we consider (1) Exact match (EM), which compares the predicted tool-using trajectory and the ground-truth ones to check if the predicted tools (names, not parameters) are exactly the same as the ground truth ones; (2) Inclusion, which measures what proportion of ground truth tools is included in the predicted tool trajectory; (3) Tool Usage (Usage), which checks if the predicted tool parameters match the ground truth ones; and (4) trajectory satisfaction (Traj-satisfy), where we prompt an LLM judge (Claude-4 by default) to determine to what extent a predicted trajectory can solve the user query, and by using this metric we mimic a real-world scenario when the ground truth is not available. For the final performance metric, we report (5) Acc, which measures if the predicted final answer matches the ground truth answer by prompting an LLM judge. For the retrieval-based methods, we include the (6) retrieval rate, which measures what proportion of ground truth tools are retrieved. Codes and data are available on \url{https://anonymous.4open.science/r/ToolData-public-2565}.

\vspace{-0.1in}
\subsection{RQ1. When and why do LLMs succeed or fail at tool-use?}\label{sec:rq1}
\vspace{-0.05in}
We first evaluate the tool-use capability of individual LLMs. For each evaluating query, we feed the LLM both the query and the available tools as the context. Since the full tool set is beyond the length of the model's context window, we focus on tools related to the same domain of the query, e.g., we only provide Travel tools for a query in the Travel domain. 
We report metrics mentioned in Section \ref{sec:eval setting} in Table \ref{tab:domain,parallel} for parallel data and Table \ref{tab:domain,sequential} for sequential data. 

\textbf{Overall tool-use performance analysis}. Generally, we observe that Gemini-2.5-pro performs best on simple queries, while Claude-4 ties with it on the hard versions. Less capable models, such as Claude-3.7 and Gemini-2.5-flash lag behind. The LLM-judge Traj-Satisfy score also tracks EM closely (e.g., Claude-4: $8.549\leftrightarrow 0.846$ on simple; $4.882\leftrightarrow 0.445$ on hard), indicating the judge is an effective proxy for EM.

In addition, we make the following key observations. (a) Most models perform well on the simple versions but struggle on the hard versions. For most metrics on most models, we observe a clear gap between simple and hard; for example, EM for Claude-4 drops from $0.846\rightarrow 0.445$ and for Gemini-2.5-pro from $0.851\rightarrow 0.442$ under direct prompting. This suggests that when tool choices and constraints must be inferred from indirect cues, 
models often miss the exact tool set and parameters, and future efforts should focus more on these more complex cases. (b) For nearly every model, \textit{parallel–simple} obtains better performance than \textit{sequential}. For example, the EM, Inclusion, and Usage of Gemini-2.5-pro ($0.851, 0.854, 0.835$) on simple parallel queries are {higher} than those on sequential queries ($0.807, 0.821, 0.809$), suggesting that inter-step dependencies and ordering introduce additional challenges for both tool selection and parameter determination. This motivates future tool-use training and inference methods that explicitly model dependency and order. (c) Inclusion typically exceeds EM, especially on hard parallel and for weaker models. For most hard cases, there is a pronounced Inclusion–EM gap—e.g., $0.135$ vs. $0.554$ for Claude-3.7 and $0.216$ vs. $0.538$ for Gemini-2.5-flash—indicating that models fail to recover the complete set. 

\begin{figure}[h!] 
\centering 
\begin{subfigure}[b]{0.45\textwidth} 
\centering \includegraphics[width=\textwidth]{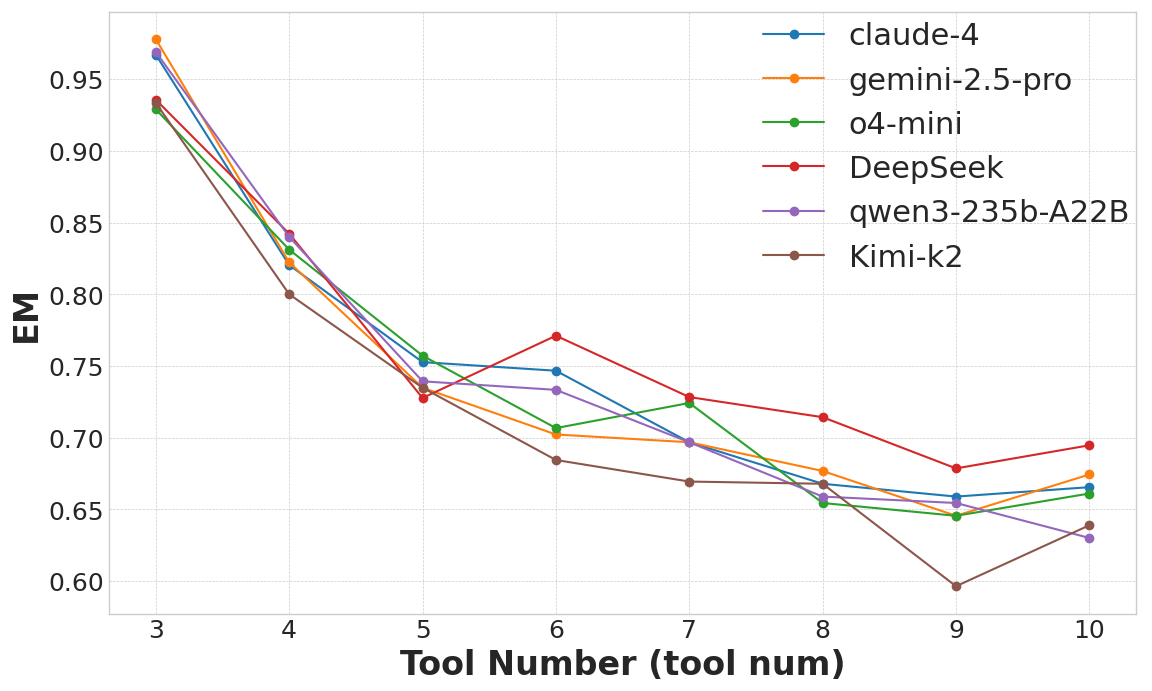}
\end{subfigure} 
\hfill 
\begin{subfigure}[b]{0.45\textwidth} 
\centering 
\includegraphics[width=\textwidth]{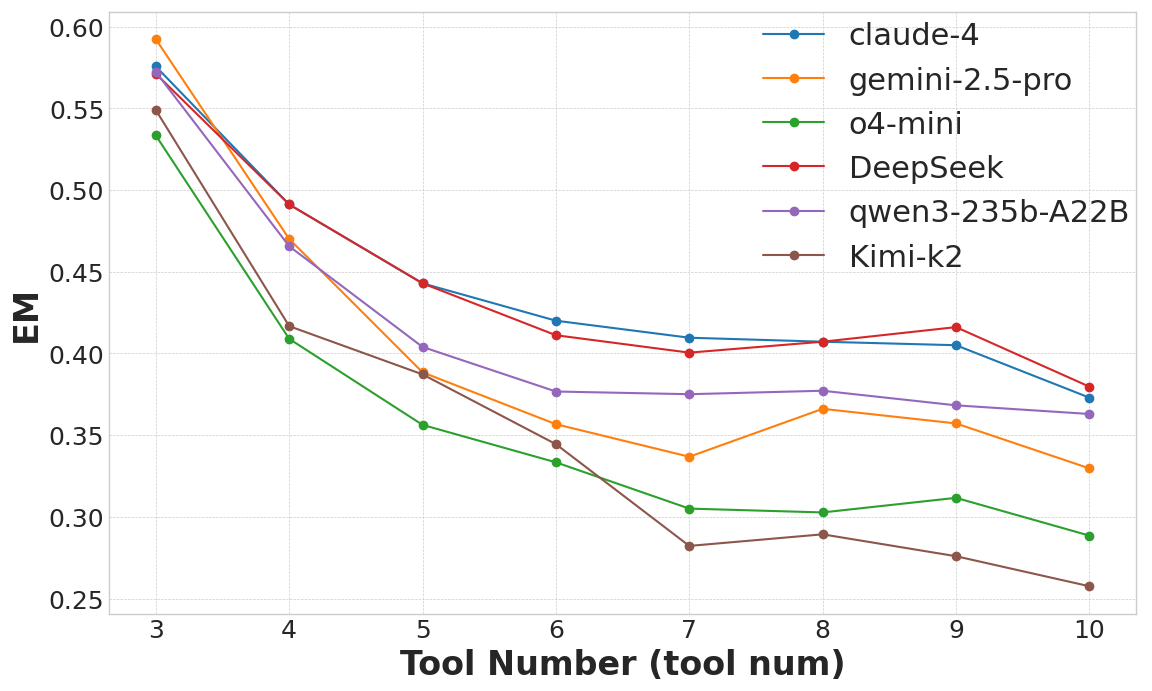} 
\end{subfigure} 
\vspace{-8pt}
\caption{\small{Figures illustrating model's scaling tool-use behavior. The x-axis denote the number of tools in the trajectory and y-axis denotes the metric EM. Left is for simple queries, while right is for hard queries.}} 
\label{fig:scaling} 
\end{figure}

\textbf{Tool-use scaling analysis}. Since the user queries can be very complex in practice and can involve many tools, we analyze the model's tool-use capability when the number of tools in the (ground truth) trajectory increases. As an illustration, we present the results on both simple and hard queries on models from different families in Figure \ref{fig:scaling}, where the x-axis is the number of tools and the y-axis is the metric EM. As the number of tools in a trajectory grows, all models experience a clear decline in performance, with the steepest drop occurring between three and five tools. 
Among the evaluated models, Claude-4 and DeepSeek show the strongest generalization, maintaining relatively stable performance even at longer tool chains, whereas o4-mini and Kimi-k2 collapse sharply beyond seven tools. These results reveal that \textbf{scaling tool use is a universal challenge}, where \textbf{the main bottleneck lies in the transition from short to mid-length trajectories}, and underline the importance of improving models’ long-horizon tool-use and error recovery capabilities. 

\textbf{Deeper analysis on tool-use failures}. We further take a look at failed cases to identify common failure patterns in tool use, and detailed examples are presented in Appendix \ref{app: failed example}. 
(a) \textbf{Similar tool confusion}. This refers to the cases when models face tools that are partially overlapping capabilities but distinct scopes, inputs/outputs, or constraints (e.g., confusing between Spotify: Search vs. YouTube Music: Search). Distinguishing among these tools requires a more delicate understanding of the tools’ exact functionalities, parameters, outputs, etc. Existing models need more improvement to obtain a precise tool match facing such distractions.
(b) \textbf{Parameter-blind tool selection \& usage}. In some cases, models overlook tool parameters (values, formats) when selecting tools, relying primarily on tool descriptions. This can lead to selecting tools with wrong parameters,cascading failures in downstream steps, and degraded end-task accuracy. We also notice that this failure is much rarer for Kimi-k2, suggesting that delicate agentic tool training can effectively improve this problem.
(c) \textbf{Redundant tool calling}. This refers to cases where the model invokes more tools than necessary. We observe two forms of redundancy: (i) \emph{related-but-unhelpful calls}—e.g., for a query about airport information in Zurich, the model additionally calls “Get All IATA airport codes,” which returns every code irrespective of the city name provided in the query; and (ii) \emph{unrelated calls}—e.g., invoking railway tools when the task asks only for flight information. The first pattern likely reflects a conservative “cover-all-bases” strategy, whereas the second is more consistent with hallucination or weak intent grounding. Since redundant calls can cause inflated latency and cost and even errors, precise tool-calling trajectory data should be included in training for improvement, and careful validation can also be adopted.
(d) \textbf{Struggle to infer intents from hard queries}. When facing indirect queries, models often misinterpret the user’s intent, leading to entirely irrelevant tool selections.

Together, these failure patterns show that models struggle with accurate tool choice and parameter use in complex scenarios, underscoring the need for better training and inference strategies.

\begin{table}[t]
\centering
\caption{\small{Results of individual model's tool-use capability with domain-specific tools on parallel queries.}}
\label{tab:domain,parallel}
\resizebox{\columnwidth}{!}{%
\begin{tabular}{l|ccccc|ccccc}
\toprule
\midrule
\multirow{2}{*}{\textbf{Model}} & \multicolumn{5}{c|}{\textbf{Simple}}                                                     & \multicolumn{5}{c}{\textbf{Hard}}                                                        \\ \cmidrule{2-11} 
                                & \textbf{EM} & \textbf{Inclusion} & \textbf{Usage} & \textbf{Traj-Satisfy} & \textbf{Acc} & \textbf{EM} & \textbf{Inclusion} & \textbf{Usage} & \textbf{Traj-Satisfy} & \textbf{Acc} \\ \midrule
\textbf{Claude-3.7}             & 0.676       & 0.746              & 0.706          & 6.831                 & 0.714        & 0.135       & 0.554              & 0.603          & 1.582                 & 0.246        \\ 
\textbf{Claude-4}               & 0.846       & 0.855              & 0.839          & 8.549                 & 0.905        & 0.445       & 0.668              & 0.794          & 4.882                 & 0.517        \\ 
\textbf{Gemini-2.5-flash}       & 0.714       & 0.759              & 0.784          & 7.215                 & 0.782        & 0.216       & 0.538              & 0.666          & 2.340                 & 0.263        \\ 
\textbf{Gemini-2.5-pro}         & 0.851       & 0.854              & 0.835          & 8.599                 & 0.911        & 0.442       & 0.652              & 0.785          & 4.849                 & 0.498        \\
\textbf{GPT5-mini}              & 0.817       & 0.825              & 0.819          & 8.236                 & 0.834        & 0.155       & 0.589              & 0.626          & 1.701                 & 0.380        \\
\textbf{o4-mini}                & 0.823       & 0.830              & 0.824          & 8.316                 & 0.863        & 0.376       & 0.629              & 0.726          & 3.849                 & 0.472        \\
\textbf{gpt-oss:120b}           & 0.652       & 0.667              & 0.715          & 6.588                 & 0.726        & 0.174       & 0.559              & 0.671          & 1.909                 & 0.387        \\
\textbf{DeepSeek}               & 0.833       & 0.836              & 0.829          & 8.417                 & 0.889        & 0.439       & 0.669              & 0.757          & 4.817                 & 0.458        \\
\textbf{qwen3-235b-A22B}        & 0.844       & 0.856              & 0.809          & 8.529                 & 0.898        & 0.440       & 0.667              & 0.796          & 4.828                 & 0.479        \\
\textbf{Kimi-k2}                & 0.815       & 0.876              & 0.872          & 8.236                 & 0.902        & 0.321       & 0.666              & 0.772          & 3.522                 & 0.448        \\ \midrule\bottomrule
\end{tabular}
}
\end{table}

\begin{table}[]
\centering
\caption{\small{Results of individual model's tool-use capability with domain-specific tools on sequential queries.}}
\label{tab:domain,sequential}
\resizebox{0.65\columnwidth}{!}{%
\begin{tabular}{l|ccccc}
\toprule
\midrule
\textbf{Model}            & \textbf{EM} & \textbf{Inclusion} & \textbf{Usage} & \textbf{Traj-Satisfy} & \textbf{Acc} \\ \midrule
\textbf{Claude-3.7}       & 0.583       & 0.724              & 0.584          & 6.010                 & 0.573        \\
\textbf{Claude-4}         & 0.819       & 0.832              & 0.775          & 8.243                 & 0.813        \\
\textbf{Gemini-2.5-flash} & 0.613       & 0.695              & 0.714          & 6.219                 & 0.652        \\
\textbf{Gemini-2.5-pro}   & 0.807       & 0.821              & 0.809          & 8.119                 & 0.848        \\
\textbf{GPT5-mini}        & 0.693       & 0.715              & 0.692          & 7.042                 & 0.677        \\
\textbf{o4-mini}          & 0.789       & 0.807              & 0.748          & 8.134                 & 0.761        \\
\textbf{gpt-oss:120b}     & 0.538       & 0.619              & 0.694          & 5.546                 & 0.653        \\
\textbf{DeepSeek}         & 0.825       & 0.849              & 0.811          & 8.305                 & 0.823        \\
\textbf{qwen3-235b-A22B}  & 0.824       & 0.837              & 0.772          & 8.194                 & 0.791        \\
\textbf{Kimi-k2}          & 0.821       & 0.827              & 0.793          & 8.363                 & 0.833        \\ \midrule\bottomrule
\end{tabular}
}
\end{table}
\vspace{-0.05in}
\subsection{RQ2. Can retrieval-augmented selection improve tool selection and parametrization?}\vspace{-0.05in}
Since tool sets can be large and diverse in practice, it is preferable to first narrow them to a smaller, more relevant subset to avoid adding irrelevant tools to the context. Retrieval-based strategy is a popular way to achieve this, and we evaluate different embedding models and different retrieval pools in experiments to find out to what extent retrieval can help.
The results in Table \ref{tab:retrieval,parallel} suggest two clear findings. (a) \textbf{When the retrieval pool is already restricted to domain-related tools, retrieval adds little benefit for simple queries}. For both Claude-3.7 and Claude-4, EM and accuracy remain nearly the same across embedding models, with ToolBench-IR showing only marginal improvements. 
(b) \textbf{Retrieval becomes a severe bottleneck for hard queries}. We notice that the retrieval rate significantly drops (merely over 50\% for most models) and all the performance metrics drop sharply compared with the non-retrieval cases. The core issue may be that retrievers heavily rely on semantic similarities and fail to capture underlying intents and steps from implicit queries. Therefore, they struggle to correctly identify necessary tools, causing cascading failures in tool selection and parameterization. 
Together, these findings reveal limitations of retrieval-based tool selection methods and call for a better strategy.

\begin{table}[]
\centering
\caption{\small{Results of individual models combined with retrieval-based tool-selection strategy.}}
\vspace{-10pt}
\label{tab:retrieval,parallel}
\resizebox{\columnwidth}{!}{%
\begin{tabular}{ll|cccccc|cccccc}
\toprule
\midrule
\multicolumn{2}{c|}{\textbf{Claude-3.7}}                  & \multicolumn{6}{c|}{\textbf{Simple}}                                       & \multicolumn{6}{c}{\textbf{Hard}}                                          \\ \midrule
\multicolumn{1}{l|}{\textbf{Emb model}}  & \textbf{Retrieval pool} & \textbf{Retrieval rate} & \textbf{EM}    & \textbf{Inclusion} & \textbf{Usage} & \textbf{Traj-Satisfy} & \textbf{Acc}   & \textbf{Retrieval rate} & \textbf{EM}    & \textbf{Inclusion} & \textbf{Usage} & \textbf{Traj-Satisfy} & \textbf{Acc}   \\ \midrule
\multicolumn{1}{l|}{\textbf{bge-large}}  & \textbf{Domain}         & 0.906          & 0.681 & 0.792     & 0.738 & 7.134        & 0.708 & 0.585          & 0.035 & 0.410     & 0.692 & 0.541        & 0.127 \\
\multicolumn{1}{l|}{}           & \textbf{All}            & 0.842          & 0.639 & 0.762     & 0.728 & 6.592        & 0.665 & 0.482          & 0.020 & 0.341     & 0.657 & 0.334        & 0.098 \\ \midrule
\multicolumn{1}{l|}{\textbf{all-MiniLM}} & \textbf{Domain}         & 0.913          & 0.685 & 0.793     & 0.749 & 7.257        & 0.717 & 0.584          & 0.029 & 0.403     & 0.683 & 0.265        & 0.109 \\
\multicolumn{1}{l|}{}           & \textbf{All}            & 0.868          & 0.645 & 0.751     & 0.745 & 6.613        & 0.680 & 0.460          & 0.012 & 0.403     & 0.641 & 0.140        & 0.082 \\ \midrule
\multicolumn{1}{l|}{\textbf{ToolLM-IR}}  & \textbf{Domain}         & 0.945          & 0.703 & 0.814     & 0.778 & 7.142        & 0.715 & 0.578          & 0.030 & 0.419     & 0.698 & 0.294        & 0.139 \\
\multicolumn{1}{l|}{}           & \textbf{All}            & 0.877          & 0.652 & 0.783     & 0.758 & 6.770        & 0.696 & 0.475          & 0.024 & 0.425     & 0.656 & 0.259        & 0.132 \\ \midrule
\multicolumn{2}{c|}{\textbf{Claude-4}}                    & \multicolumn{6}{c|}{\textbf{Simple}}                              & \multicolumn{6}{c}{\textbf{Hard}}                                 \\ \midrule
\multicolumn{1}{l|}{\textbf{bge-large}}  & \textbf{Domain}         & 0.906          & 0.852 & 0.867     & 0.835 & 8.631        & 0.902 & 0.585          & 0.031 & 0.397     & 0.672 & 0.328        & 0.264 \\
\multicolumn{1}{l|}{}           & \textbf{All}            & 0.842          & 0.785 & 0.823     & 0.770 & 8.053        & 0.876 & 0.482          & 0.012 & 0.292     & 0.656 & 0.111        & 0.189 \\ \midrule
\multicolumn{1}{l|}{\textbf{all-MiniLM}} & \textbf{Domain}         & 0.913          & 0.833 & 0.859     & 0.786 & 8.764        & 0.870 & 0.584          & 0.033 & 0.410     & 0.663 & 0.355        & 0.277 \\
\multicolumn{1}{l|}{}           & \textbf{All}            & 0.868          & 0.817 & 0.806     & 0.773 & 8.352        & 0.832 & 0.460          & 0.015 & 0.267     & 0.620 & 0.171        & 0.168 \\ \midrule
\multicolumn{1}{l|}{\textbf{ToolLM-IR}}  & \textbf{Domain}         & 0.945          & 0.906 & 0.928     & 0.833 & 9.117        & 0.916 & 0.578          & 0.028 & 0.420     & 0.680 & 0.286        & 0.241 \\
\multicolumn{1}{l|}{}           & \textbf{All}            & 0.877          & 0.852 & 0.861     & 0.764 & 8.613        & 0.879 & 0.475          & 0.014 & 0.298     & 0.653 & 0.190        & 0.164 \\ \midrule\bottomrule
\end{tabular}
}
\end{table}

\begin{table}[]
\centering
\caption{\small{Evaluation of model's inherent agentic tool-use capability, denoted as ``agentic". ``context" denotes the results providing tools as context, same with Table \ref{tab:domain,parallel}.}}
\vspace{-5pt}
\label{tab:model agentic,parallel}
\resizebox{\columnwidth}{!}{%
\begin{tabular}{l|l|ccccc|ccccc}
\toprule
\midrule
\multirow{2}{*}{\textbf{model}}          & \multirow{2}{*}{} & \multicolumn{5}{c|}{\textbf{Simple}}                      & \multicolumn{5}{c}{\textbf{Hard}}                         \\ \cmidrule{3-12} 
                                         &                   & \textbf{EM}    & \textbf{Inclusion} & \textbf{Usage} & \textbf{Traj-Satisfy} & \textbf{Acc}   & \textbf{EM}    & \textbf{Inclusion} & \textbf{Usage} & \textbf{Traj-Satisfy} & \textbf{Acc}   \\ \midrule
\multirow{2}{*}{\textbf{Claude-4}}       & \textbf{Agentic}           & 0.832 & 0.868     & 0.816 & 8.407        & 0.893 & 0.440 & 0.637     & 0.751 & 4.828        & 0.486 \\
                                         & \textbf{Context}           & 0.846 & 0.855     & 0.839 & 8.549        & 0.905 & 0.445 & 0.668     & 0.794 & 4.882        & 0.517 \\ \midrule
\multirow{2}{*}{\textbf{Gemini-2.5-pro}} & \textbf{Agentic}           & 0.828 & 0.866     & 0.876 & 8.367        & 0.917 & 0.416 & 0.674     & 0.768 & 4.564        & 0.503 \\
                                         & \textbf{Context}           & 0.851 & 0.847     & 0.835 & 8.599        & 0.911 & 0.442 & 0.652     & 0.785 & 4.849        & 0.498 \\ \midrule
\multirow{2}{*}{\textbf{Deepseek}}       & \textbf{Agentic}           & 0.819 & 0.857     & 0.756 & 8.256        & 0.845 & 0.416 & 0.682     & 0.783 & 4.564        & 0.481 \\
                                         & \textbf{Context}           & 0.835 & 0.833     & 0.729 & 8.417        & 0.889 & 0.439 & 0.669     & 0.757 & 4.817        & 0.458 \\ \midrule
\multirow{2}{*}{\textbf{Kimi-k2}}        & \textbf{Agentic}           & 0.853 & 0.977     & 0.893 & 8.620        & 0.951 & 0.315 & 0.717     & 0.765 & 3.456        & 0.437 \\
                                         & \textbf{Context}           & 0.815 & 0.876     & 0.872 & 8.236        & 0.902 & 0.321 & 0.666     & 0.772 & 3.522        & 0.448 \\ \midrule\bottomrule
\end{tabular}
}
\end{table}
\vspace{-0.05in}
\subsection{RQ3. Do agentic methods, including training and inference, improve tool-use capability?}\vspace{-0.05in}
Despite the individual model's capability of using tool information as the context, agentic methods are also developed in existing literature, including two major categories: one is the agentic tool-use training undergone by most SOTA models \citep{team2025kimi, yang2025qwen3}, and the other is to design LLM agents to adopt tools \citep{yao2023react}. For the first category, we evaluate four representative models that inherently support tool-use, and show results in Table \ref{tab:model agentic,parallel}, where we also provide results of context-based evaluation as a baseline. Table \ref{tab:model agentic,parallel} shows that the agentic tool-use capability is similar to the context-based baseline for most models on both simple and hard queries. 

For the second category, we evaluate Claude models on both parallel and sequential queries by combining retrieval with ReAct. We consider two settings: a static mode, where a subset of tools is retrieved once based on the user query, and a dynamic mode, where retrieval occurs before each thought and action. ToolLM-IR is used as the retrieval backbone in all experiments. The results in Tables \ref{tab:react,parallel} and \ref{tab:react,sequential} show that ReAct consistently improves tool-use performance compared with individual models in Table \ref{tab:domain,parallel}. For example, on parallel hard queries, Claude-4 improves from 0.445 EM (Table \ref{tab:domain,parallel}) to 0.463 EM with ReAct (Table \ref{tab:react,parallel}), and the performance is further boosted via dynamic retrieval to 0.473 EM. Similarly, Claude-3.7 gains from 0.135 EM to 0.186 EM with domain tools and 0.296 EM under dynamic ReAct. These show that \textbf{iterative calling tools based on execution results offer a stronger basis for accurate tool retrieval and usage}.

Overall, these findings suggest that agentic tool learning and inference serve as an effective and robust way to improve a model's tool-use capability.

\begin{table}[h]
\centering

\vspace{0.05in}
\caption{\small{ReAct results on parallel queries combined with different retrieval strategy.}}
\vspace{-5pt}
\label{tab:react,parallel}
\resizebox{\columnwidth}{!}{%
\begin{tabular}{l|l|ccccc|ccccc}
\toprule
\midrule
\textbf{}                            & \textbf{}               & \multicolumn{5}{c|}{\textbf{Simple}}                                                     & \multicolumn{5}{c}{\textbf{Hard}}                                                        \\ \midrule
\textbf{Model}                       & \textbf{retrieval mode} & \textbf{EM} & \textbf{Inclusion} & \textbf{Usage} & \textbf{Traj-Satisfy} & \textbf{Acc} & \textbf{EM} & \textbf{Inclusion} & \textbf{Usage} & \textbf{Traj-Satisfy} & \textbf{Acc} \\ \midrule
\multirow{3}{*}{\textbf{claude-3.7}} & \textbf{Domain tool}    & 0.735       & 0.816              & 0.752          & 7.630                 & 0.782        & 0.186       & 0.612              & 0.641          & 1.930                 & 0.307        \\
                                     & \textbf{Static}         & 0.762       & 0.841              & 0.791          & 7.840                 & 0.803        & 0.093       & 0.432              & 0.697          & 1.180                 & 0.148        \\
                                     & \textbf{Dynamic}        & 0.814       & 0.876              & 0.805          & 8.220                 & 0.831        & 0.296       & 0.581              & 0.703          & 3.080                 & 0.346        \\ \midrule
\multirow{3}{*}{\textbf{claude-4}}   & \textbf{Domain tool}    & 0.892       & 0.915              & 0.864          & 9.160                 & 0.928        & 0.463       & 0.670              & 0.797          & 4.804                 & 0.418        \\
                                     & \textbf{Static}         & 0.886       & 0.912              & 0.872          & 9.116                 & 0.916        & 0.031       & 0.424              & 0.668          & 0.403                 & 0.139        \\
                                     & \textbf{Dynamic}        & 0.933       & 0.951              & 0.875          & 9.422                 & 0.948        & 0.472       & 0.703              & 0.736          & 4.811                 & 0.438        \\ \midrule\bottomrule
\end{tabular}
}
\end{table}

\begin{table}[h]
\centering
\vspace{0.05in}
\caption{\small{ReAct results on sequential queries combined with different retrieval strategy.}}
\vspace{-5pt}
\label{tab:react,sequential}
\resizebox{0.8\columnwidth}{!}{%
\begin{tabular}{l|l|ccccc}
\toprule
\midrule
\textbf{Sequential}                  & \textbf{Retrieval Mode} & \textbf{EM} & \textbf{Inclusion} & \textbf{Usage} & \textbf{Traj-Satisfy} & \textbf{Acc} \\ \midrule
\multirow{3}{*}{\textbf{claude-3.7}} & \textbf{Domain tool}    & 0.615       & 0.735              & 0.625          & 6.384                 & 0.634        \\
                                     & \textbf{Static}         & 0.609       & 0.692              & 0.644          & 6.266                 & 0.603        \\
                                     & \textbf{Dynamic}        & 0.651       & 0.749              & 0.651          & 6.574                 & 0.634        \\ \midrule
\multirow{3}{*}{\textbf{claude-4}}   & \textbf{Domain tool}    & 0.827       & 0.846              & 0.829          & 8.492                 & 0.842        \\
                                     & \textbf{Static}         & 0.817       & 0.833              & 0.825          & 8.059                 & 0.811        \\
                                     & \textbf{Dynamic}        & 0.849       & 0.881              & 0.833          & 8.573                 & 0.917        \\ \midrule\bottomrule
\end{tabular}
}
\end{table}

\vspace{-0.05in}
\section{Conclusion}
\vspace{-0.05in}
This work introduces a comprehensive tool-use benchmark that focuses on the tool-use trajectory. By constructing a large and diverse executable tool suite, modeling trajectories of varying structures and scales, and pairing them with user queries of different difficulty levels, we provide a realistic and rigorous setting for evaluation. Our trajectory-aware metrics go beyond final-answer accuracy, enabling a clearer understanding of where and why LLMs succeed or fail in tool use. Through comprehensive evaluation and analysis, our benchmark not only highlights the current strengths and limitations of state-of-the-art models but also offers actionable insights for improving agentic tool-use capabilities. We hope this benchmark establishes a foundation for systematic progress in developing LLMs that can reliably plan, select, and execute tools in complex, real-world scenarios.

\section*{Ethics statement}
We acknowledge the ICLR Code of Ethics and ensure that no concerns regarding the Code of Ethics arise from our work. Our study does not involve human subjects, personal or sensitive data, or experiments that could cause harm. All data used are either synthetic or publicly available under appropriate licenses, and we have adhered to principles of fairness, transparency, and reproducibility throughout.

\section*{Reproducibility statement}
We provide data and codes in \url{https://anonymous.4open.science/r/ToolData-public-2565}, mentioned in section \ref{sec:eval setting}. We provide additional details in Appendix \ref{app: exp detail}.

\bibliography{iclr2026_conference}
\bibliographystyle{iclr2026_conference}

\appendix
\section{\bench~details}
We provide more details of the proposed \bench.
\subsection{tool set details and examples}\label{app:tool details}
\textbf{Overall statictics}. We provide statistics of the tool set in \bench~in Table \ref{tab:tool stat}. 
\begin{table}[h]
  \centering
  \caption{Statistics of tool set in \bench}
  \label{tab:tool stat}
\resizebox{0.3\columnwidth}{!}{
\begin{tabular}{c|c}
\toprule
\midrule
\textbf{domain}      & \textbf{tool num} \\ \midrule
\textbf{Travel}      & 131               \\
\textbf{Mapping}     & 124               \\
\textbf{Finance}     & 258               \\
\textbf{Weather}     & 135               \\
\textbf{eCommerce}   & 82                \\
\textbf{Mews\_Media} & 118               \\
\textbf{Gaming}      & 84                \\
\textbf{Email}       & 70                \\
\textbf{Education}   & 89                \\
\textbf{Music}       & 137               \\ \midrule
\textbf{Total}       & 1228              \\ \midrule\bottomrule
\end{tabular}
}
\end{table}

\textbf{Tool examples}. We then provide some examples of tools in \bench~ as follows.
\begin{lstlisting}[basicstyle=\ttfamily\small,breaklines=true,frame=single]
{
  "tool name": "Airbnb listings: Prices and Availability by lat/lng",
  "tool description": "Retrieves average prices, availability percentages, and processed property counts within a 20,000-meter radius of a geographical point, filterable by bedrooms and guest capacity. This analytics endpoint leverages the parent tool's access to daily-updated Airbnb data, providing insights into property availability and pricing trends to help users make informed decisions about accommodations.",
  "required_parameters": [
    { "name": "year", "type": "NUMBER", "description": "the year", "default": 2024 },
    { "name": "lat", "type": "NUMBER", "description": "latitude", "default": 28.0852473 },
    { "name": "lng", "type": "NUMBER", "description": "longitude", "default": -16.7349705 },
    { "name": "range", "type": "NUMBER", "description": "range in meters from latitude and longitude point", "default": 500 },
    { "name": "month", "type": "STRING", "description": "the month", "default": "1" }
  ],
  "optional_parameters": [
    { "name": "bedrooms", "type": "NUMBER", "description": "number of bedrooms", "default": 1 },
    { "name": "maxguestcapacity", "type": "NUMBER", "description": "maximum guest capacity", "default": 2 }
  ],
  "API name": "Prices and Availability by lat/lng",
  "domain name": "Travel",
  "output_info": {
    "output_info_summary": "The tool provides a statistical summary of Airbnb listings within a specified geographical radius. A successful response returns aggregated data, including the total number of properties, median prices for all and for available properties, and the overall availability percentage. This information is intended to help users understand local accommodation market trends. The tool also returns a simple error message object in cases of failure, such as an authentication or subscription issue.",
    "output_format": "The output is a single, flat JSON object. The structure differs for successful and failed requests. A successful response contains keys like `totalProperties`, `medianPrice`, and `availability`. An error response typically contains a single `message` key with a string value. There is no indication of nesting, arrays, or pagination."
  },
  "connected tools": [
    {
      "tool name": "Airbnb listings: Listings by lat/lng",
      "connect params": [
        { "name": "lat", "description": "Can use the same latitude from the source tool's query" },
        { "name": "lng", "description": "Can use the same longitude from the source tool's query" },
        { "name": "range", "description": "Can use the same or adjusted range based on property density from source tool" }
      ]
    },
    {
      "tool name": "Priceline com Provider: Search hotels locations by geolocation",
      "connect params": [
        { "name": "latitude", "description": "Can use the same latitude from the source tool's query" },
        { "name": "longitude", "description": "Can use the same longitude from the source tool's query" }
      ]
    }
  ]
},
{
    "tool name": "Billboard: Hot 100",
    "tool description": "Provides detailed information about Billboard's Hot 100 chart, displaying the current rankings of the most popular songs in the United States. This endpoint delivers access to Billboard's comprehensive music chart system, which includes not only the Hot 100 but also other major charts like Billboard 200, Billboard Global 200, and Artist 100. Billboard charts are the music industry's standard for measuring the popularity of artists, albums, and songs based on sales, radio airplay, streaming data, and other metrics, offering authoritative insights into current music trends and performance.",
    "required_parameters": [
      {
        "name": "date",
        "type": "DATE (YYYY-MM-DD)",
        "description": "Set the date for which you want to get a chart.",
        "default": "2020-03-18"
      }
    ],
    "optional_parameters": [],
    "API name": "Hot 100",
    "domain name": "Music",
    "output_info": {
      "output_info_summary": "The tool returns a ranked list of songs from the Billboard Hot 100 chart. For each song, it provides details such as the artist, title, current rank, last week's rank, peak position, and the number of weeks it has been on the chart. This data allows users to track the current popularity and historical performance of top songs in the United States.",
      "output_format": "The output is a string-serialized JSON array of objects. Each object corresponds to a song and its chart data. The 'artist' field contains HTML `<a>` tags linking to Billboard's website. The 'weeks_on_chart' field can be an empty string, likely for new entries. Some string values may contain HTML character entities (e.g., `&#039;`)."
    },
    "connected tools": [
      {
        "tool name": "Spotify: Search",
        "connect params": [
          {
            "name": "q",
            "description": "The song title and artist from Hot 100 can be used as search query"
          },
          {
            "name": "type",
            "description": "Can be set to 'tracks' to search for songs from the Hot 100 chart"
          }
        ]
      },
      {
        "tool name": "Shazam: search",
        "connect params": [
          {
            "name": "term",
            "description": "The song title and artist from Hot 100 can be used as search term"
          }
        ]
      },
      {
        "tool name": "Youtube Music API (Detailed): Search",
        "connect params": [
          {
            "name": "query",
            "description": "The song title and artist from Hot 100 can be used as search query"
          }
        ]
      }
    ]
  }
\end{lstlisting}

\textbf{Execution errors of tools}. As we mentioned in Section \ref{sec:data:tool} 

\subsection{Trajectory and query details}\label{app:traj details}
We provide more details of tool-use trajectories and queries in \bench.

\textbf{Trajectory statistics}. The statistics of queries and trajectories in \bench~ are shown in Table \ref{tab:query stat}.

\begin{table}[h]
\centering
  \caption{Statistics of queries and trajectories in \bench}
  \label{tab:query stat}
\resizebox{0.6\columnwidth}{!}{
\begin{tabular}{c|ccc}
\toprule
\midrule
\textbf{domain}      & \textbf{simple parallel} & \textbf{hard parallel} & \textbf{sequential} \\ \midrule
\textbf{Travel}      & 200                      & 200                    & 170                 \\
\textbf{Mapping}     & 200                      & 200                    & 200                 \\
\textbf{Finance}     & 200                      & 200                    & 135                 \\
\textbf{Weather}     & 200                      & 200                    & 200                 \\
\textbf{eCommerce}   & 200                      & 200                    & 200                 \\
\textbf{Mews\_Media} & 200                      & 200                    & 180                 \\
\textbf{Gaming}      & 200                      & 200                    & 185                 \\
\textbf{Email}       & 200                      & 200                    & 200                 \\
\textbf{Education}   & 200                      & 200                    & 200                 \\
\textbf{Music}       & 200                      & 200                    & 200                 \\ \midrule
\textbf{Total}       & 2000                     & 2000                   & 1870                \\ \midrule\bottomrule
\end{tabular}
}
\end{table}

\textbf{Examples of task types for parallel queries}. We provide some examples of task types used in parallel query generations.

\begin{lstlisting}[basicstyle=\ttfamily\small,breaklines=true,frame=single]
{
    "domain": "Travel",
    "task name": "Find Proper Accommodations",
    "task description": "Find proper hotels based on some options.  It requires searching across different platforms, comparing options based on real-time data, and retrieving specific, granular details like policies, recent reviews, or room availability that are crucial for making a booking decision.",
    "tool classes": [
        "Hotel",
        "Airbnb"
    ],
    "task examples": [
        {
            "query": "I'm planning a trip to London from Oct 10 to Oct 17 in 2025. I need to find the best hotel deal. Can you search on both Booking.com and Hotels.com? I want a hotel with at least a 4-star rating and I need to see its cancellation policy and some recent photos before I decide.",
            "tool list": [
                "Best Booking.com Hotel: find best booking.com accommodation",
                "Hotels com Provider: Hotels Search",
                "Booking.com_v2: stays/properties/detail/policies",
                "Booking.com_v2: stays/properties/detail/photos"
            ]
        }
    ]
},
{
    "domain": "Gaming",
    "task name": "Tabletop & Strategy Game Assistance",
    "task description": "This task focuses on the non-video game tools, providing a user with rule-based assistance for a traditional tabletop or card game, such as calculating odds or determining the mathematically optimal move.",
    "tool classes": [
        "Card_Dice_Games",
        "Chess_Puzzles"
    ],
    "task examples": [
        {
            "query": "I'm playing a game of Blackjack and need some help. The dealer is showing a 7. My hand is two 8s. According to basic strategy, what is the optimal move? Also, I think the deck is rich in high cards; the running count is +8 and there are about 4 decks left in the shoe. What is the True Count?",
            "tool list": [
                "BlackJack Basic Strategy: Pairs",
                "BlackJack Basic Strategy: True Count"
            ]
        }
    ]
},
{
    "domain": "Finance",
    "task name": "In-Depth Company Analysis",
    "task description": "This task involves creating a holistic financial overview of a specific company/stock by combining its reported fundamentals, current market data, and the consensus opinion of professional analysts.",
    "tool classes": [
        "Fundamental_Data",
        "Analyst_Ratings_Estimates"
    ],
    "task examples": [
        {
            "query": "Give me a complete financial workup on Apple (AAPL). I want to see their latest quarterly income statement and balance sheet. Also, pull the current institutional ownership percentage and any recent insider transactions. Finally, what is the consensus analyst price target for the stock?",
            "tool list": [
                "Financial Modeling Prep: Company Income Statement",
                "Financial Modeling Prep: Companies balance sheet statement",
                "Mboum Finance: stock/institution-ownership",
                "EOD Historical Data: Insider Transactions API",
                "YFINANCE - Stock Market Data: Analyst Price Target"
            ]
        }
    ]
}
\end{lstlisting}

\textbf{Examples of task sequences for sequential queries}. We also provide some examples of task sequences used to generate exact sequential queries. It is clear that we define a coherence chain for each task and specify connections (parameter used for subsequent tools) among tools. Note that these sequences are general and do not contain any exact information such as detailed parameter values, so they can serve as templates to create exact trajectories and queries fitting different realistic scenarios.

\begin{lstlisting}[basicstyle=\ttfamily\small,breaklines=true,frame=single]
{
    "domain": "Music",
    "name": "Radio Genre to Artist Profile Discovery",
    "description": "This sequence starts by discovering radio stations in a specific country to identify a music genre, then finds an artist in that genre on Spotify, and concludes by retrieving that artist's detailed profile.",
    "tool sequence": [
    {
    "tool name": "50K Radio Stations: Get Countries",
    "use description": "Retrieves a list of countries with radio stations to select a country ID.",
    "param for next tool": "country_id"
    },
    {
    "tool name": "50K Radio Stations: Get Channels",
    "use description": "Finds radio channels in the selected country to identify a prominent genre.",
    "param for next tool": "q"
    },
    {
    "tool name": "Spotify: Search",
    "use description": "Searches for artists on Spotify using the genre as a query to get an artist's ID.",
    "param for next tool": "artist_id"
    },
    {
    "tool name": "Spotify _v2: Artist Details (Single)",
    "use description": "Fetches the complete profile for the artist from Spotify using their ID.",
    "param for next tool": ""
    }
    ],
    "tool number": 4
},
{
    "domain": "Education",
    "name": "DEI Resource -- Speaker Analysis -- Vocabulary Deep Dive -- Further Study",
    "description": "This sequence models a student's workflow: finding a DEI resource, identifying a speaker, finding their TED talks, and then performing a deep vocabulary analysis on a key term before exploring related learning resources.",
    "tool sequence": [
    {
    "tool name": "DEI: GetLearningResources",
    "use description": "A user gets a list of DEI learning resources and selects one, which provides a 'title'.",
    "param for next tool": "title"
    },
    {
    "tool name": "DEI: GetPeople",
    "use description": "The 'title' is used to find an influential person related to that topic, providing their 'name'.",
    "param for next tool": "name"
    },
    {
    "tool name": "TED Talks API: getTalks",
    "use description": "The person's 'name' is used as the 'speaker' to find their TED talks, which returns a 'talk_title'.",
    "param for next tool": "talk_title"
    },
    {
    "tool name": "Urban Dictionary: Define a Word / Search",
    "use description": "A key term from the 'talk_title' is used as the 'term' to look up its modern, colloquial, or slang meaning.",
    "param for next tool": "term"
    },
    {
    "tool name": "Dictionary: wordSearchTurkish",
    "use description": "The same 'term' is used as a 'query' to find its Turkish translation, providing an 'anlam' (meaning).",
    "param for next tool": "anlam"
    },
    {
    "tool name": "Urban Dictionary: Define a Word / Search",
    "use description": "The Turkish 'anlam' (meaning) is used as a 'term' in a reverse-lookup to see its English slang definitions.",
    "param for next tool": "topic"
    },
    {
    "tool name": "DEI: GetLearningResources",
    "use description": "Finally, the original 'topic' from the talk is used to find other, related learning resources.",
    "param for next tool": ""
    }
    ],
    "tool number": 7
}
\end{lstlisting}

\textbf{Examples of queries and trajectories}. We then provide examples of trajectories and queries in \bench.

\textit{Parallel}.
\begin{lstlisting}[basicstyle=\ttfamily\small,breaklines=true,frame=single, escapeinside={(*@}{@*)}]
(*@\textbf{Example 1: Travel domain}@*)
(*@\textcolor{green}{Tool Trajectory}@*): [
      {
        "tool name": "Priceline com Provider: Hotel reviews",
        "tool description": "Returns a list of reviews for travel services such as hotels, cars, and flights. This endpoint is part of the Priceline.com API, which enables users to search and book travel accommodations and rentals. It leverages a database of user-generated reviews to provide insights, helping travelers make informed decisions about their bookings.",
        "required parameters": [
          {
            "name": "hotel_id",
            "value": "700022612"
          }
        ],
        "optional parameters": [
          {
            "name": "languages",
            "value": "en,fr"
          },
          {
            "name": "limit",
            "value": 100
          },
          {
            "name": "only_verified_guests",
            "value": true
          }
        ],
        "executed_output": "{'getHotelReviews': {'error': {'status': 'Hotel.Reviews: No reviews found as Offset value exceeds the number of reviews.', 'status_code': '1.822.8', 'time': '0.0715'}}}"
      },
      {
        "tool name": "Booking.com_v2: languages",
        "tool description": "Retrieves available language options for hotel information and booking interfaces. This endpoint leverages Booking.com's unofficial API to access real-time data from hotels worldwide, allowing users to specify their preferred language when querying room availability, pricing, facilities, and policies as part of comprehensive travel planning.",
        "required parameters": [],
        "optional parameters": [],
        "executed_output": "{'data': [{'__ref': 'Language:{\"code\":\"en-gb\"}'}, {'__ref': 'Language:{\"code\":\"en-us\"}'}, {'__ref': 'Language:{\"code\":\"de\"}'}, {'__ref': 'Language:{\"code\":\"nl\"}'}, {'__ref': 'Language:{\"code\":\"fr\"}'}, {'__ref': 'Language:{\"code\":\"es\"}'}, {'__ref': 'Language:{\"code\":\"es-ar\"}'}, {'__ref': 'Language:{\"code\":\"es-mx\"}'}, {'__ref': 'Language:{\"code\":\"ca\"}'}, {'__ref': 'Language:{\"code\":\"it\"}'}, {'__ref': 'Language:{\"code\":\"pt-pt\"}'}, {'__ref': 'Language:{\"code\":\"pt-br\"}'}, {'__ref': 'Language:{\"code\":\"no\"}'}, {'__ref': 'Language:{\"code\":\"fi\"}'}, {'__ref': 'Language:{\"code\":\"sv\"}'}, {'__ref': 'Language:{\"code\":\"da\"}'}, {'__ref': 'Language:{\"code\":\"cs\"}'}, {'__ref': 'Language:{\"code\":\"hu\"}'}, {'__ref': 'Language:{\"code\":\"ro\"}'}, {'__ref': 'Language:{\"code\":\"ja\"}'}, {'__ref': 'Language:{\"code\":\"zh-cn\"}'}, {'__ref': 'Language:{\"code\":\"zh-tw\"}'}, {'__ref': 'Language:{\"code\":\"pl\"}'}, {'__ref': 'Language:{\"code\":\"el\"}'}, {'__ref': 'Language:{\"code\":\"ru\"}'}, {'__ref': 'Language:{\"code\":\"tr\"}'}, {'__ref': 'Language:{\"code\":\"bg\"}'}, {'__ref': 'Language:{\"code\":\"ar\"}'}, {'__ref': 'Language:{\"code\":\"ko\"}'}, {'__ref': 'Language:{\"code\":\"he\"}'}, {'__ref': 'Language:{\"code\":\"lv\"}'}, {'__ref': 'Language:{\"code\":\"uk\"}'}, {'__ref': 'Language:{\"code\":\"hi\"}'}, {'__ref': 'Language:{\"code\":\"id\"}'}, {'__ref': 'Language:{\"code\":\"ms\"}'}, {'__ref': 'Language:{\"code\":\"th\"}'}, {'__ref': 'Language:{\"code\":\"et\"}'}, {'__ref': 'Language:{\"code\":\"hr\"}'}, {'__ref': 'Language:{\"code\":\"lt\"}'}, {'__ref': 'Language:{\"code\":\"sk\"}'}, {'__ref': 'Language:{\"code\":\"sr\"}'}, {'__ref': 'Language:{\"code\":\"sl\"}'}, {'__ref': 'Language:{\"code\":\"vi\"}'}, {'__ref': 'Language:{\"code\":\"tl\"}'}, {'__ref': 'Language:{\"code\":\"is\"}'}], 'message': 'Successful', 'status': True}"
      },
      {
        "tool name": "Cities Cost of Living: Get Cities List",
        "tool description": "Retrieves a comprehensive list of all available cities in the database. This endpoint taps into a global dataset covering over 650 cities worldwide, providing access to the complete catalog of locations for which detailed living expense information is available. The parent service offers extensive cost-of-living data that helps users compare and understand financial requirements across different urban centers globally.",
        "required parameters": [],
        "optional parameters": [],
        "executed_output": "{'cities': [{'country': 'Turkey', 'name': 'Kocaeli'}, {'country': 'Czech Republic', 'name': 'Ostrava'}, {'country': 'United States', 'name': 'Santa Clara'}, {'country': 'United States', 'name': 'Iowa City'}, {'country': 'United Kingdom', 'name': 'Bournemouth'}, {'country': 'Montenegro', 'name': 'Podgorica'}, {'country': 'United States', 'name': 'Toledo'}, {'country': 'United Kingdom', 'name': 'Milton Keynes'}, {'country': 'India', 'name': 'Mangalore'}, {'country': 'Malaysia', 'name': 'Johor Bahru'}, {'country': 'Canada', 'name': \"St. John's\"}, {'country': 'Pakistan', 'name': 'Islamabad'}, {'country': 'Norway', 'name': 'Trondheim'}, {'country': 'India', 'name': 'Nagpur'}, {'country': 'Czech Republic', 'name': 'Olomouc'}, {'country': 'Netherlands', 'name': 'Best'}, {'country': 'Ukraine', 'name': 'Zaporizhzhya'}, {'country': 'India', 'name': 'Mysore'}, {'country': 'China', 'name': 'Beijing'}, {'country': 'North Macedonia', 'name': 'Skopje'}, {'country': 'Philippines', 'name': 'Manila'}, {'country': 'Netherlands', 'name': 'Rotterdam'}, {'country': 'Taiwan (China)', 'name': 'Taipei'}, {'country': 'Brazil', 'name': 'Goiania'}, {'country': 'Taiwan (China)', 'name': 'New Taipei City'}, {'country': 'Austria', 'name': 'Salzburg'}, {'country': 'Germany', 'name': 'Frankfurt'}, {'country': 'Turkey', 'name': 'Diyarbakir'}, {'country': 'Canada', 'name': 'North Vancouver'}, {'country': 'India', 'name': 'Ludhiana'}, {'country': 'Israel', 'name': 'Jerusalem'}, {'country': 'Afghanistan', 'name': 'Kabul'}, {'country': 'Romania', 'name': 'Iasi'}, {'country': 'United Kingdom', 'name': 'Bristol'}, {'country': 'South Africa', 'name': 'Port Elizabeth'}, {'country': 'Ecuador', 'name': 'Guayaquil'}, {'country': 'France', 'name': 'Bordeaux'}, {'country': 'Georgia', 'name': 'Batumi'}, {'country': 'Botswana', 'name': 'Gaborone'}, {'country': 'Brazil', 'name': 'Natal'}, {'country': 'Trinidad And Tobago', 'name': 'Port of Spain'}, {'country': 'Belgium', 'name': 'Antwerp'}, {'country': 'Austria', 'name': 'Vienna'}, {'country': 'France', 'name': "
      }
    ],
    "trajectory_type": "parallel",
    "tool count": 3,
    "final_answer": "# Travel Logistics Information\n\n## 1. Hotel Reviews from Priceline.com\n\nThe Priceline.com Provider tool returned an error when attempting to retrieve hotel reviews. The specific error message was:\n\n\"Hotel.Reviews: No reviews found as Offset value exceeds the number of reviews.\"\n\nThis suggests that either the hotel ID provided (700022612) doesn't have any reviews, or there was an issue with the parameters used in the request. To get valid hotel reviews, you would need to provide a different hotel ID or adjust the offset parameter.\n\n## 2. Languages Supported by Booking.com\n\nBooking.com supports 45 languages across various regions. The complete list includes:\n\n- English (UK): en-gb\n- English (US): en-us\n- German: de\n- Dutch: nl\n- French: fr\n- Spanish: es\n- Spanish (Argentina): es-ar\n- Spanish (Mexico): es-mx\n- Catalan: ca\n- Italian: it\n- Portuguese (Portugal): pt-pt\n- Portuguese (Brazil): pt-br\n- Norwegian: no\n- Finnish: fi\n- Swedish: sv\n- Danish: da\n- Czech: cs\n- Hungarian: hu\n- Romanian: ro\n- Japanese: ja\n- Chinese (Simplified): zh-cn\n- Chinese (Traditional): zh-tw\n- Polish: pl\n- Greek: el\n- Russian: ru\n- Turkish: tr\n- Bulgarian: bg\n- Arabic: ar\n- Korean: ko\n- Hebrew: he\n- Latvian: lv\n- Ukrainian: uk\n- Hindi: hi\n- Indonesian: id\n- Malay: ms\n- Thai: th\n- Estonian: et\n- Croatian: hr\n- Lithuanian: lt\n- Slovak: sk\n- Serbian: sr\n- Slovenian: sl\n- Vietnamese: vi\n- Tagalog: tl\n- Icelandic: is\n\n## 3. Cities for Cost-of-Living Comparisons\n\nThe Cities Cost of Living tool returned a partial list of cities available for cost-of-living comparisons. The output was truncated, showing only the first 43 cities out of what appears to be a much larger database (reportedly covering over 650 cities worldwide). \n\nSome of the cities included in the partial list:\n- Kocaeli, Turkey\n- Ostrava, Czech Republic\n- Santa Clara, United States\n- Iowa City, United States\n- Bournemouth, United Kingdom\n- Beijing, China\n- Manila, Philippines\n- Frankfurt, Germany\n- Vienna, Austria\n\nPlease note that this is only a small subset of the complete city list, as the tool output was cut off after 43 cities.\n\n# Summary\n\nI've gathered information on languages supported by Booking.com (45 languages) and a partial list of cities available for cost-of-living comparisons. Unfortunately, the hotel reviews request returned an error, so no review data is available with the provided parameters. To complete your travel logistics preparation, you may need to request hotel reviews with different parameters and obtain the complete city list for cost-of-living comparisons."
}
(*@\textcolor{blue}{Simple query}@*): I'm preparing travel logistics. Can you grab: Hotel reviews from Priceline.com; which languages Booking.com supports; and the city list for cost of living comparisons?
(*@\textcolor{red}{Hard query}@*): I'm pulling together travel logistics for a fall trip. Could you first scan recent guest feedback from a major U.S. hotel aggregator and summarize consistent pros/cons for central options? Next, check which languages the leading global booking portal actually supports across its site and app (not just auto-translate), and note any gaps. Finally, put together a short list of major cities we can use for side-by-side cost-of-living comparisons.

(*@\textbf{Example 2: eCommerce domain}@*)
(*@\textcolor{green}{Tool Trajectory}@*): [
      {
        "tool name": "Wayfair: auto-complete",
        "tool description": "Retrieves auto suggestions based on a provided term or phrase, helping users quickly find relevant content. This API provides comprehensive search functionality for Wayfair's extensive catalog, allowing you to query for information about categories, products, and other offerings just as you would on wayfair.com. The service streamlines the discovery process by suggesting relevant terms as users type, making it easier to navigate through Wayfair's extensive home goods and furniture inventory.",
        "required parameters": [
          {
            "name": "query",
            "value": "red office chair"
          }
        ],
        "optional parameters": [],
        "executed_output": "{'summary': {'request_id': '', 'transaction_id': 'u4aD+FSTT4S/epB1PGO3bw==', 'page_type': 'General', 'response_hash': '231be424e42fe70df57916c01134581d1986d7a4', 'response_matches_prior_hash': False, 'cache_seconds': 1800, 'cache_always_check_server': False, 'spv_custom_vars': 'LoginStatusFlag=0'}, 'response': [{'schema_id': 'WFSearchSuggestion', 'value': 'red office chair', 'type': 'keyword', 'first_in_section': False, 'is_reform': False, 'keyword': False}, {'schema_id': 'WFSearchSuggestion', 'value': 'office chair red', 'type': 'keyword', 'first_in_section': False, 'is_reform': False, 'keyword': False}, {'schema_id': 'WFSearchSuggestion', 'value': 'red office chairs with gold frames', 'type': 'keyword', 'first_in_section': False, 'is_reform': False, 'keyword': False}]}"
      },
      {
        "tool name": "Asos: v2/auto-complete",
        "tool description": "Gets autocomplete suggestions based on a partial product name input, helping users quickly find specific items they're looking for. This endpoint leverages the Asos API's comprehensive product database to deliver relevant search completions, functioning just like the autocomplete feature on the official Asos website. The API provides access to the same extensive catalog of categories, products, and related information that powers the Asos shopping platform.",
        "required parameters": [
          {
            "name": "q",
            "value": "bikini top"
          }
        ],
        "optional parameters": [
          {
            "name": "store",
            "value": "US"
          },
          {
            "name": "country",
            "value": "US"
          },
          {
            "name": "currency",
            "value": "USD"
          },
          {
            "name": "lang",
            "value": "en-US"
          }
        ],
        "executed_output": "{'suggestionGroups': [{'indexName': 'searchterms', 'indexTitle': 'searchterms', 'suggestions': [{'searchTerm': 'bikini top', 'numberOfResults': 2460}, {'searchTerm': 'underwire bikini top', 'numberOfResults': 213}, {'searchTerm': 'bandeau bikini top', 'numberOfResults': 225}, {'searchTerm': 'black bikini top', 'numberOfResults': 414}, {'searchTerm': 'fuller bust bikini top', 'numberOfResults': 95}, {'searchTerm': 'white bikini top', 'numberOfResults': 303}, {'searchTerm': 'triangle bikini top', 'numberOfResults': 519}]}]}"
      },
      {
        "tool name": "Asos: categories/list",
        "tool description": "Lists all available product categories from Asos, providing a structured overview of the shopping taxonomy. This endpoint taps into the comprehensive Asos API system that mirrors the official website's data architecture, allowing users to efficiently navigate through the complete category hierarchy before diving into specific product searches or filtering options.",
        "required parameters": [],
        "optional parameters": [
          {
            "name": "lang",
            "value": "en-US"
          },
          {
            "name": "country",
            "value": "US"
          }
        ],
        "executed_output": "{'navigation': [{'id': '7276d7f9-b810-4743-8c11-eccb260bbecd', 'alias': 'MW', 'type': 'link', 'channelExclusions': [], 'webLargePriority': 0, 'content': {'title': 'Men', 'subTitle': None, 'webLargeImageUrl': None, 'mobileImageUrl': None}, 'display': None, 'style': {'webLargeStyleType': 'dark', 'mobileStyleType': 'dark'}, 'link': {'linkType': 'internal', 'brandSectionAlias': None, 'categoryId': None, 'webUrl': 'https://www.asos.com/us/men/', 'appUrl': None}, 'children': [{'id': 'ae28af2b-e3ca-4f2f-a559-9a976a0812d4', 'alias': None, 'type': 'container', 'channelExclusions': [], 'webLargePriority': 0, 'content': {'title': 'Home', 'subTitle': None, 'webLargeImageUrl': None, 'mobileImageUrl': None}, 'display': {'webLargeTemplateId': 0, 'webLargeTemplateName': '', 'webLargeColumnSpan': 1, 'mobileTemplateId': 7, 'mobileTemplateName': 'homepageLinks', 'mobileDisplayLayout': 'list'}, 'style': {'webLargeStyleType': 'dark', 'mobileStyleType': 'noTitle'}, 'link': None, 'children': [{'id': '47989db5-0597-43e6-9bee-1348c35590b0', 'alias': None, 'type': 'link', 'channelExclusions': ['apps', 'webLarge'], 'webLargePriority': 0, 'content': {'title': 'Home', 'subTitle': None, 'webLargeImageUrl': 'https://images.asos-media.com/navigation/mw_homebuttonnew_1826147', 'mobileImageUrl': 'https://images.asos-media.com/navigation/mw_homebuttonnew_1826147'}, 'display': None, 'style': {'webLargeStyleType': 'dark', 'mobileStyleType': 'dark'}, 'link': {'linkType': 'internal', 'brandSectionAlias': None, 'categoryId': None, 'webUrl': 'https://www.asos.com/us/men/', 'appUrl': None}, 'children': []}]}, {'id': 'da3b6d82-c0e8-423f-a6ed-f23e930ce951', 'alias': None, 'type': 'container', 'channelExclusions': ['webLarge'], 'webLargePriority': 0, 'content': {'title': 'App and Mobile Top Level - Carousel', 'subTitle': None, 'webLargeImageUrl': None, 'mobileImageUrl': None}, 'display': {'webLargeTemplateId': 0, 'webLargeTemplateName': '', 'webLargeColumnSpan': 1, 'mobileTemplateId': 1, 'mobileTemplateName': 'imageBackground', 'mobileDisplayLayout': 'list'"
      }
    ],
    "trajectory_type": "parallel",
    "tool count": 3,
    "final_answer": "# Retailer Comparison: Wayfair and Asos\n\n## Wayfair Search Results for 'red office chair'\nI used Wayfair's auto-complete tool to search for 'red office chair' and found these suggested search terms:\n- red office chair\n- office chair red\n- red office chairs with gold frames\n\nThe tool only provided search suggestions and did not return actual product listings, prices, or availability information.\n\n## Asos Search Results for 'bikini top'\nI searched Asos for 'bikini top' using the US store, US country setting, USD currency, and English language. The auto-complete tool returned these suggestions with their respective product counts:\n- bikini top (2,460 products)\n- underwire bikini top (213 products)\n- bandeau bikini top (225 products)\n- black bikini top (414 products)\n- fuller bust bikini top (95 products)\n- white bikini top (303 products)\n- triangle bikini top (519 products)\n\nThis indicates Asos has a substantial selection of bikini tops with various styles available.\n\n## Asos Categories\nI attempted to retrieve the full category list from Asos for the US market, but the tool returned only partial information. The output shows the beginning of the men's category structure but was cut off. The data includes:\n- Main category: Men\n- Some navigation structure information\n- Partial subcategory information\n\nThe tool did not return the complete category list, particularly missing women's categories which would be most relevant for bikini tops.\n\n## Summary\nThis comparison provides a limited snapshot of both retailers. For Wayfair, we only have search suggestions for red office chairs without actual product information. For Asos, we have quantitative data showing they offer over 2,400 bikini top products with various style options, but the category listing was incomplete. To make a comprehensive comparison between these retailers, additional information about actual products, pricing, shipping options, and return policies would be needed."
}
(*@\textcolor{blue}{Simple query}@*): I'm comparing retailers and want a clear snapshot. Please get Wayfair auto-complete suggestions for 'red office chair', and get Asos auto-complete suggestions for 'bikini top' with store US, country US, currency USD, lang en-US, and list Asos categories with country US, lang en-US.
(*@\textcolor{red}{Hard query}@*):We're moving next month and I'm trying to stretch a tight budget without buying junk. Here are the references I've jotted down: On Wayfair, I'm thinking along the lines of 'red office chair' show what's trending. On Asos since I'm shopping from the US and paying in dollars, I'm exploring 'bikini top' to round out accessories. I want to see how Asos organizes things for a US selection in English so I don't miss a section. If something is clearly better value, flag it\u2014otherwise show me the top few comparable picks.
\end{lstlisting}

\textit{Sequential}.
\begin{lstlisting}[basicstyle=\ttfamily\small,breaklines=true,frame=single, escapeinside={(*@}{@*)}]
(*@\textcolor{blue}{Query}@*): "I'm looking for Ed Sheeran's \"Shape of You\" on SoundCloud. Can you find the track, then get detailed metadata for it using the official SoundCloud URL, and finally download the timed lyrics by matching it with the third candidate on Spotify? I want the lyrics for track 301161123 since it has over 12,000 comments and is one of his most popular songs from 2017.",
(*@\textcolor{green}{Tool trajectory}@*): [
      {
        "tool name": "Miza: Song search",
        "tool description": "Searches for a song on SoundCloud using the 'scsearch:' prefix to get its direct URL.",
        "required parameters": [
          {
            "name": "search",
            "value": "scsearch:Shape of You Ed Sheeran"
          }
        ],
        "optional parameters": [],
        "execution_status": "success",
        "executed_output": "[{'name': 'Shape of You', 'url': 'https://soundcloud.com/edsheeran/shape-of-you', 'duration': 233.759, 'icon': 'https://i1.sndcdn.com/artworks-jnr3tXcz4dKQ-0-original.jpg'}]",
        "API name": "Song search",
        "domain name": "Music",
        "parent tool name": "Miza",
        "sequence_step": {
          "step_number": 1,
          "tool_name": "Miza: Song search",
          "description": "Searches for a song on SoundCloud using the 'scsearch:' prefix to get its direct URL.",
          "param_for_next_tool": "track",
          "original_description": "{'tool name': 'Miza: Song search', 'use description': \"Searches for a song on SoundCloud using the 'scsearch:' prefix to get its direct URL.\", 'param for next tool': 'track'}"
        },
        "original_description": "{'tool name': 'Miza: Song search', 'use description': \"Searches for a song on SoundCloud using the 'scsearch:' prefix to get its direct URL.\", 'param for next tool': 'track'}"
      },
      {
        "tool name": "SoundCloud Scraper: Get Track Metadata (1-3 Quotas)",
        "tool description": "Retrieves the full metadata for the SoundCloud track using its URL.",
        "required parameters": [
          {
            "name": "track",
            "value": "https://soundcloud.com/edsheeran/shape-of-you"
          }
        ],
        "optional parameters": [],
        "execution_status": "success",
        "executed_output": "{'status': True, 'errorId': 'Success', 'audio': [{'quality': 'sq', 'url': 'https://scd.dlod.link/', 'durationMs': 233744, 'durationText': '03:53', 'mimeType': 'audio/mpeg', 'extension': 'mp3'}, {'quality': 'sq', 'url': 'https://scd.dlod.link/', 'durationMs': 233719, 'durationText': '03:53', 'mimeType': 'audio/ogg; codecs=\"opus\"', 'extension': 'opus'}], 'type': 'track', 'id': 301161123, 'permalink': 'https://soundcloud.com/edsheeran/shape-of-you', 'createdAt': '2017-01-06T04:05:41Z', 'lastModified': '2025-09-05T06:03:33Z', 'title': 'Shape of You', 'caption': None, 'description': None, 'goPlus': False, 'durationMs': 233744, 'durationText': '03:53', 'artworkUrl': 'https://i1.sndcdn.com/artworks-cK0SqfJHnCqU-0-original.jpg', 'releaseDate': '2017-01-06T00:00:00Z', 'genre': 'Pop', 'labelName': 'Atlantic Records UK', 'license': 'all-rights-reserved', 'purchaseTitle': None, 'purchaseUrl': None, 'commentable': True, 'commentCount': 12127, 'likeCount",
        "API name": "Get Track Metadata (1-3 Quotas)",
        "domain name": "Music",
        "parent tool name": "SoundCloud Scraper",
        "sequence_step": {
          "step_number": 2,
          "tool_name": "SoundCloud Scraper: Get Track Metadata (1-3 Quotas)",
          "description": "Retrieves the full metadata for the SoundCloud track using its URL.",
          "param_for_next_tool": "track",
          "original_description": "{'tool name': 'SoundCloud Scraper: Get Track Metadata (1-3 Quotas)', 'use description': 'Retrieves the full metadata for the SoundCloud track using its URL.', 'param for next tool': 'track'}"
        },
        "original_description": "{'tool name': 'SoundCloud Scraper: Get Track Metadata (1-3 Quotas)', 'use description': 'Retrieves the full metadata for the SoundCloud track using its URL.', 'param for next tool': 'track'}",
        "adapt_parameter": "track",
        "adapt_constraint": "track URL for a popular Ed Sheeran song",
        "adapt_reason": "Selected the URL for 'Shape of You' by Ed Sheeran as it represents the track identifier in the SoundCloud ecosystem. This URL uniquely identifies the track and is the most appropriate parameter for the SoundCloud Scraper's 'track' parameter."
      },
      {
        "tool name": "SoundCloud Scraper: Download Lyrics on Spotify (2 Quotas)",
        "tool description": "Uses the SoundCloud track URL to find a match on Spotify and download its timed lyrics.",
        "required parameters": [
          {
            "name": "track",
            "value": "301161123"
          }
        ],
        "optional parameters": [
          {
            "name": "candidate",
            "value": 3
          }
        ],
        "execution_status": "success",
        "executed_output": "[00:09.77]The club isn't the best place to find a lover\n[00:12.16]So the bar is where I go, mm-mm\n[00:14.94]Me and my friends at the table doin' shots\n[00:17.16]Drinkin' fast and then we talk slow, mm-mm\n[00:19.41]And you come over and start up a conversation with just me\n[00:22.36]And trust me, I'll give it a chance now, mm-mm\n[00:24.68]Take my hand, stop, put Van the Man on the jukebox\n[00:27.36]And then we start to dance, and now I'm singin' like\n[00:29.95]Girl, you know I want your love\n[00:32.43]Your love was handmade for somebody like me\n[00:35.55]Come on now, follow my lead\n[00:37.40]I may be crazy, don't mind me\n[00:39.58]Say, \"Boy, let's not talk too much\"\n[00:42.44]Grab on my waist and put that body on me\n[00:45.59]Come on now, follow my lead\n[00:47.13]Come, come on now, follow my lead, mm-mm\n[00:50.88]I'm in love with the shape of you\n[00:53.21]We push and pull like a magnet do\n[00:55.76]Although my heart is fallin' too\n[00:58.35]I'm in love with your body\n[01:00.66]And last night, you were in my room\n[01:03.12]And now my bedsheets smell like you\n[01:05.35]Every day, discoverin' somethin' brand-new\n[01:08.13]Well, I'm in love with your body\n[01:10.02]Oh, I, oh, I, oh, I, oh, I\n[01:13.34]Well, I'm in love with your body\n[01:14.91]Oh, I, oh, I, oh, I, oh, I\n[01:18.20]Well, I'm in love with your body\n[01:19.92]Oh, I, oh, I, oh, I, oh, I\n[01:23.22]Well, I'm in love with your body\n[01:25.33]Every day, discoverin' somethin' brand-new\n[01:28.39]I'm in love with the shape ",
        "API name": "Download Lyrics on Spotify (2 Quotas)",
        "domain name": "Music",
        "parent tool name": "SoundCloud Scraper",
        "sequence_step": {
          "step_number": 3,
          "tool_name": "SoundCloud Scraper: Download Lyrics on Spotify (2 Quotas)",
          "description": "Uses the SoundCloud track URL to find a match on Spotify and download its timed lyrics.",
          "param_for_next_tool": "",
          "original_description": "{'tool name': 'SoundCloud Scraper: Download Lyrics on Spotify (2 Quotas)', 'use description': 'Uses the SoundCloud track URL to find a match on Spotify and download its timed lyrics.', 'param for next tool': ''}"
        },
        "original_description": "{'tool name': 'SoundCloud Scraper: Download Lyrics on Spotify (2 Quotas)', 'use description': 'Uses the SoundCloud track URL to find a match on Spotify and download its timed lyrics.', 'param for next tool': ''}",
        "adapt_parameter": "track",
        "adapt_constraint": "track ID for a popular song with high engagement",
        "adapt_reason": "Selected 'Shape of You' by Ed Sheeran (ID: 301161123) because it has significant engagement (12127+ comments) and is a well-known track from 2017. The track ID is the unique identifier needed for the SoundCloud Scraper."
      }
    ],
    "domain": "Music",
    "final_answer": "I successfully found Ed Sheeran's 'Shape of You' on SoundCloud and retrieved the detailed metadata and timed lyrics as requested. Here's what I found:\n\n1. SoundCloud Search Results:\n   - Successfully located 'Shape of You' by Ed Sheeran\n   - URL: https://soundcloud.com/edsheeran/shape-of-you\n   - Duration: 233.759 seconds (approximately 3:53)\n\n2. Track Metadata from SoundCloud:\n   - Confirmed this is track ID 301161123 as you mentioned\n   - The track has two audio quality options available (mp3 and opus formats)\n   - Duration matches the search result at approximately 3:53\n\n3. Timed Lyrics:\n   - Successfully retrieved the synchronized lyrics from Spotify (third candidate match as requested)\n   - The lyrics are time-stamped, starting at [00:09.77] with \"The club isn't the best place to find a lover\"\n   - The lyrics continue throughout the song with precise timestamps\n\nThe track you specified (301161123) was indeed found and appears to be the official version from Ed Sheeran's SoundCloud account, matching your description as one of his most popular songs from 2017."
\end{lstlisting}

\section{Additional evaluation details}\label{app: exp detail}

In this section we provide more details about evaluation.

\textbf{Evaluation on individual LLMs}. For this evaluation, we provide both test query and available tools as context to the model and prompt the model to predict a trajectory. Specifically, we enforce the model to given answer in JSON format. The prompt is as follows. 
\begin{lstlisting}[basicstyle=\ttfamily\small,breaklines=true,frame=single, escapeinside={(*@}{@*)}]
(*@\textbf{Direct Prompt}@*)
Given the tool list:
<tools>

Please slove the query with help of these tools if necessary:
<query>

Please selct proper tools and, provide me proper parameters to call it. If you need outputs from previous tools, use a placeholder '<results from tool xxx>' .Please respond in the following json format:
```json
{
\"tool list\": [
{\"tool name\":[tool name], 
\"tool description\":[tool description], 
\"required_parameters\":[{\"name\": xxx,\"value\": xxx}, {\"name\": xxx,\"value\": xxx},...], 
\"optional_parameters\":[{\"name\": xxx,\"value\": xxx},...], 
\"API name\":[API name], 
\"domain name\":[domain name], 
\"parent tool name\":[parent tool name]},
...
]
}
```
\end{lstlisting}

Despite the direct prompting, we also include CoT prompting.
\begin{lstlisting}[basicstyle=\ttfamily\small,breaklines=true,frame=single, escapeinside={(*@}{@*)}]
(*@\textbf{CoT prompt}@*)
You are a problem solver.
Goal: given a user query and a pool of tools, produce the MINIMAL, CORRECT tool-call trajectory that satisfies the query.
Please think step by step to:
1) extract intent, entities, constraints, required outputs,
2) shortlist feasible tools (match inputs\u2192outputs),
3) design a trajectory (sequential or parallel),
4) compute arguments with correct types/formats/units,
5) check dependencies and stop criteria.

Here are some examples:
Query: Schedule dinner with Alice at 19:00\u201321:00, 2025-8-19 in San Jose at a vegan place; invite alice@example.com.
Tools: [\n  {\n    \"tool name\": \"Calendar: create_calendar_event\",\n    \"tool description\": \"Create an event; emails participants.\",\n    \"required_parameters\": [\n    {\n    \"name\": \"title\",\n    \"type\": \"string\",\n    \"default\": \"set up a meeting\",\n    \"description\": \"The title of the event\",\n    },\n    {\n    \"name\": \"start_time\",\n    \"type\": \"string\",\n    \"default\": \"2025-8-19 19:00\",\n    \"description\": \"The start time of the event\",\n    },\n    {\n    \"name\": \"end_time\",\n    \"type\": \"string\",\n    \"default\": \"2025-8-19 21:00\",\n    \"description\": \"The end time of the event\",\n    },\n    {\n    \"name\": \"participants\",\n    \"type\": \"list<email>\",\n    \"default\": [\"alice@example.com\"],\n    \"description\": \"The participants of the event\",\n    }\n    ],\n    \"optional_parameters\": [\n    {\n    \"name\": \"location\",\n    \"type\": \"str\",\n    \"default\": \"San Jose\",\n    \"description\": \"The location of the event\",\n    }\n    ],\n    \"API name\": \"create_calendar_event\",\n    \"domain name\": \"Travel\",\n    \"parent tool name\": \"Calendar\"\n  },\n  {\n    \"tool name\": \"Restaurant: get_all_restaurants_in_city\",\n    \"tool description\": \"List restaurant names in a city.\",\n    \"required_parameters\": [\n    {\n    \"name\": \"city\",\n    \"type\": \"str\",\n    \"default\": \"San Jose\",\n    \"description\": \"The city of the event\",\n    }\n    ],\n    \"optional_parameters\": [],\n    \"API name\": \"get_all_restaurants_in_city\",\n    \"domain name\": \"Travel\",\n    \"parent tool name\": \"Restaurant\"\n  },\n  {\n    \"tool name\": \"Restaurant: check_restaurant_opening_hours\",\n    \"tool description\": \"Return operating hours; use to verify time.\",\n    \"required_parameters\": [\n    {\n    \"name\": \"restaurant_names\",\n    \"type\": \"list<str>\",\n    \"default\": [\"restaurant1\", \"restaurant2\"],\n    \"description\": \"The names of the restaurants\",\n    }\n    ],\n    \"optional_parameters\": [],\n    \"API name\": \"check_restaurant_opening_hours\",\n    \"domain name\": \"Travel\",\n    \"parent tool name\": \"Restaurant\"\n  },\n  ...\n]
thought: Let's think step by step. First I need to find a vegan restuarant in San Jose. I need to call tool Restaurant: get_all_restaurants_in_city with city=San Jose. After I identify the vegan restaurant, I need to check if the restaurant is open that time. I need to call tool Restaurant: check_restaurant_opening_hours with restaurant_names=[\"vegan restaurant\"]. Finally, I need to set up this dinner on my calendar and send to Alice. I need to call tool Calendar: create_calendar_event with title=\"Dinner with Alice\", start_time=\"2025-8-19 19:00\", end_time=\"2025-8-19 21:00\", participants=[\"alice@example.com\"], location=\"San Jose\". 
tool list: [\n{\"tool name\":\"Restaurant: get_all_restaurants_in_city\", \"tool description\": \"List restaurant names in a city.\", \"required_parameters\":[{\"name\": \"city\", \"value\": \"San Jose\"}], \"optional_parameters\":[], \"API name\":\"get_all_restaurants_in_city\", \"domain name\":\"Travel\", \"parent tool name\":\"Restaurant\"},\n...\n]
...
\end{lstlisting}

For the models, we use their default/recommended temperatures and allow thinking if it is a reasoning model like Claude/Gemini/...

\textbf{Retrieval-base selection}. By default, we retrieve 20 tools per time, and feed them to the model as the context.

\section{Additional experiments}\label{app:exp}

We present additional results of evaluating individual models with CoT prompting in Table \ref{tab:model, parallel, cot} and \ref{tab:model, sequential, cot}. It is clear that CoT does not bring much improvement compared with the direct prompting. This may be due to the fact that almost all the evaluated models are reasoning models and we already allow the thinking mode by default. This suggest that simple direct prompting is already good enough, which is consistent with conclusions  in \citep{guo2025deepseekr1}.

\begin{table}[h]
\centering
\caption{Results of individual LLMs using CoT on parallel queries.}
\label{tab:model, parallel, cot}
\resizebox{\columnwidth}{!}{%
\begin{tabular}{l|ccccc|ccccc}
\toprule
\midrule
\multirow{2}{*}{\textbf{Model}} & \multicolumn{5}{c|}{\textbf{Simple}}                                                     & \multicolumn{5}{c}{\textbf{Hard}}                                                        \\ \cmidrule{2-11} 
                                & \textbf{EM} & \textbf{Inclusion} & \textbf{Usage} & \textbf{Traj-Satisfy} & \textbf{Acc} & \textbf{EM} & \textbf{Inclusion} & \textbf{Usage} & \textbf{Traj-Satisfy} & \textbf{Acc} \\ \midrule
\textbf{Claude-3.7}             & 0.635       & 0.663              & 0.698          & 6.318                 & 0.709        & 0.106       & 0.383              & 0.603          & 1.163                 & 0.243        \\
\textbf{Claude-4}               & 0.826       & 0.839              & 0.799          & 8.218                 & 0.882        & 0.438       & 0.636              & 0.801          & 4.460                 & 0.492        \\
\textbf{Gemini-2.5-flash}       & 0.698       & 0.719              & 0.813          & 6.945                 & 0.768        & 0.203       & 0.491              & 0.692          & 2.227                 & 0.250        \\
\textbf{Gemini-2.5-pro}         & 0.774       & 0.793              & 0.735          & 7.831                 & 0.867        & 0.393       & 0.628              & 0.695          & 4.119                 & 0.451        \\
\textbf{GPT5-mini}              & 0.782       & 0.801              & 0.804          & 7.911                 & 0.795        & 0.168       & 0.524              & 0.681          & 1.843                 & 0.369        \\
\textbf{o4-mini}                & 0.771       & 0.762              & 0.821          & 7.671                 & 0.842        & 0.344       & 0.614              & 0.732          & 3.774                 & 0.464        \\
\textbf{gpt-oss:120b}           & 0.629       & 0.645              & 0.712          & 6.358                 & 0.713        & 0.158       & 0.523              & 0.657          & 1.734                 & 0.355        \\
\textbf{DeepSeek}               & 0.832       & 0.839              & 0.818          & 8.278                 & 0.875        & 0.425       & 0.661              & 0.762          & 4.663                 & 0.431        \\
\textbf{qwen3-235b-A22B}        & 0.807       & 0.815              & 0.789          & 8.029                 & 0.863        & 0.364       & 0.651              & 0.813          & 3.994                 & 0.463        \\
\textbf{Kimi-k2}                & 0.783       & 0.838              & 0.869          & 7.891                 & 0.875        & 0.328       & 0.642              & 0.757          & 3.599                 & 0.416        \\ \midrule\bottomrule
\end{tabular}
}
\end{table}

\begin{table}[h]
\centering
\caption{Results of individual LLMs using CoT on sequential queries.}
\label{tab:model, sequential, cot}
\resizebox{0.75\columnwidth}{!}{%
\begin{tabular}{l|ccccc}
\toprule
\midrule
\textbf{Model}            & \textbf{EM} & \textbf{Inclusion} & \textbf{Usage} & \textbf{Traj-Satisfy} & \textbf{Acc} \\ \midrule
\textbf{Claude-3.7}       & 0.570       & 0.650              & 0.615          & 5.950                 & 0.612        \\
\textbf{Claude-4}         & 0.805       & 0.811              & 0.767          & 8.103                 & 0.818        \\
\textbf{Gemini-2.5-flash} & 0.598       & 0.691              & 0.693          & 6.023                 & 0.629        \\
\textbf{Gemini-2.5-pro}   & 0.786       & 0.792              & 0.786          & 7.920                 & 0.825        \\
\textbf{GPT5-mini}        & 0.687       & 0.711              & 0.704          & 6.913                 & 0.648        \\
\textbf{o4-mini}          & 0.764       & 0.793              & 0.754          & 7.975                 & 0.774        \\
\textbf{gpt-oss:120b}     & 0.497       & 0.593              & 0.679          & 5.088                 & 0.625        \\
\textbf{DeepSeek}         & 0.818       & 0.836              & 0.805          & 8.239                 & 0.798        \\
\textbf{qwen3-235b-A22B}  & 0.786       & 0.805              & 0.784          & 7.905                 & 0.757        \\
\textbf{Kimi-k2}          & 0.793       & 0.814              & 0.791          & 8.078                 & 0.825        \\ \midrule\bottomrule
\end{tabular}
}
\end{table}

\section{Failed examples}\label{app: failed example}
We present examples corresponding to failure patterns discussed in section \ref{sec:rq1}.

\textit{Similar tool confusion}.
\begin{lstlisting}[basicstyle=\ttfamily\small,breaklines=true,frame=single, escapeinside={(*@}{@*)}]
(*@\textbf{Query}@*): Could you help me collect: hotel amenity codes list (limit 100); us states gas price; Airbnb stays near 25.7617,-80.1918 within ~10 km?
(*@\textcolor{red}{Incorrect tool list}@*):['Airbnb listings: Listings by lat lng', 'Gas Price: stateUsaPrice', 'Airbnb listings: Amenities']
(*@\textcolor{blue}{Correct tool list}@*): ['Priceline com Provider: Download filter amenities', 'Gas Price: stateUsaPrice', 'Airbnb listings: Listings by lat lng']
\end{lstlisting}
The model wrongly select 'Airbnb listings: Amenities' (mixed with 'Priceline com Provider: Download filter amenities') while the query asks for hotel amenity.
\begin{lstlisting}[basicstyle=\ttfamily\small,breaklines=true,frame=single, escapeinside={(*@}{@*)}]
(*@\textbf{Query}@*): Share the current weather there in metric, a short-term nowcast for that spot, and a quick read on recent air pollution?
(*@\textcolor{red}{Incorrect tool list}@*):['AI Weather by Meteosource: current', 'Foreca Weather: Nowcast', 'RapidWeather: Current air pollution data']
(*@\textcolor{blue}{Correct tool list}@*): ['AI Weather by Meteosource: current', 'Foreca Weather: Nowcast', 'RapidWeather: Historical air pollution data']
\end{lstlisting}
The model wrongly select 'RapidWeather: Current air pollution data' (mixed with 'RapidWeather: Historical air pollution data') while the intent is to retrieve historical data.

\textit{Parameter-blind tool selection and use}.
\begin{lstlisting}[basicstyle=\ttfamily\small,breaklines=true,frame=single, escapeinside={(*@}{@*)}]
(*@\textbf{Query}@*): ... followed by looking up airports in France, ...
(*@\textcolor{red}{Incorrect tool}@*):{'tool name': 'Flight Data_v2: Airport data in json format',  'tool description': 'Returns a file containing a comprehensive list of airports from the database. This endpoint is part of the Travelpayouts Data API, which provides valuable travel insights for websites and blogs by offering access to flight price trends and popular destination data that can help you better serve your customers with relevant travel information.',  'required parameters': [],  'optional parameters': [],...}
(*@\textcolor{blue}{Correct tool}@*): {'tool name': 'Flightera Flight Data: airportSearch',  'tool description': "Retrieves a list of airports based on country, bounding box, or timezone, allowing multiple parameters for precise filtering. This endpoint is part of Flightera's comprehensive service for flight status, on-time performance, and statistics, leveraging a database of 60k airports to ensure accurate and actionable data for flight planning and real-time analytics.",  'required parameters': [],  'optional parameters': [{'name': 'country', 'value': 'FR'}],...}
\end{lstlisting}
The model wrongly select a tool, which can retrieve airport data but does not have correct parameter (France).

\textit{Redundant tool calling}.

Related but not helpful.
\begin{lstlisting}[basicstyle=\ttfamily\small,breaklines=true,frame=single, escapeinside={(*@}{@*)}]
(*@\textbf{Query}@*): What travel information is available for Switzerland? I'm interested in comparing options in Zurich and Geneva, including hotel availability and airport information.
(*@\textcolor{red}{Incorrect tool}@*):{'tool name': 'iata_airport_codes: Get All iata airport codes',  'tool description': 'Retrieves a comprehensive list of all IATA airport codes, sourced from a globally maintained database. This endpoint provides access to the core dataset of the parent tool, which serves as an authoritative reference for airport code information, enabling applications to validate locations, integrate travel data, or analyze aviation networks with accurate, standardized identifiers.',  'required parameters': [],  'optional parameters': []}
\end{lstlisting}

Unrelated tools.
\begin{lstlisting}[basicstyle=\ttfamily\small,breaklines=true,frame=single, escapeinside={(*@}{@*)}]
(*@\textbf{Query}@*): I am planning a trip to Switzerland. I'm interested in comparing options in Zurich and Geneva, including hotel availability and airport information,...
(*@\textcolor{red}{Incorrect tool}@*):{'tool name': 'SBB Suisse railway: Autocomplete',  'tool description': "Searches for train and public transport stations in Switzerland and across Europe, integrated with the SBB's railway network and public transport data to provide detailed journey planning, including fare calculations, departure and arrival times, and route options between any two stops in Switzerland.",  'required parameters': [{'name': 'query', 'value': 'Geneva'}],  'optional parameters': []}
\end{lstlisting}

\textit{Fail to infer underlying requirements}
\begin{lstlisting}[basicstyle=\ttfamily\small,breaklines=true,frame=single, escapeinside={(*@}{@*)}]
(*@\textbf{Query}@*): I am planning a trip with my family. I'd prefer somewhere that past guests have loved ...
(*@\textcolor{red}{Incorrect tool}@*):{'tool name': 'Priceline com Provider: Search hotels locations',  'tool description': "Searches for travel locations by name, allowing you to find specific destinations for your trip planning. This endpoint is part of priceline.com's comprehensive travel booking service that lets you search for hotels, rental cars, and flights across multiple destinations. The service provides access to priceline.com's extensive database of travel options, helping you quickly locate and compare travel destinations to build your ideal itinerary.",  'required parameters': [{'name': 'name', 'value': 'Vancouver'},  {'name': 'search_type', 'value': 'ALL'}],  'optional parameters': [],...}
(*@\textcolor{blue}{Correct tool}@*): {'tool name': 'Hotels com Provider: Hotels Search',  'tool description': 'Searches for hotels based on location, dates, and other criteria, providing comprehensive information about available accommodations including prices, amenities, and reviews.',  'required_parameters': [{'name': 'checkin_date', 'value': '2024-07-01'},  {'name': 'checkout_date', 'value': '2024-07-08'},  {'name': 'sort_order', 'value': 'REVIEW'}
\end{lstlisting}
The model fails to interpret 'somewhere that past guests have loved' into 'somewhere with good customer reviews', and therefore does not incorporate such parameters.

\end{document}